\documentclass[11pt]{article}

\usepackage[final]{acl}

\usepackage{times}
\usepackage{latexsym}

\usepackage[T1]{fontenc}

\usepackage[utf8]{inputenc}

\usepackage{microtype}

\usepackage{inconsolata}

\usepackage{graphicx}
\usepackage{verbatim}
\usepackage{amsmath}
\usepackage{multirow}
\usepackage{multicol}
\usepackage{booktabs}
\usepackage{enumitem}
\usepackage{array}
\usepackage{xcolor}
\usepackage{float}
\usepackage{algorithm}       
\usepackage{algorithmic}
\usepackage{amssymb}
\usepackage{tabularx}
\usepackage{threeparttable}
\usepackage{subcaption}  
\usepackage{colortbl}
\usepackage{tikz}
\usepackage{longtable}
\usepackage{pifont}
\usepackage{tcolorbox}

\newcommand{\cmark}{\ding{51}}
\newcommand{\xmark}{\ding{55}}
\newcommand{\corpusname}{\textsc{Indica}}
\newcommand{\sel}{\textcolor{green!60!black}{$\checkmark$} }
\definecolor{lightearthygreen}{HTML}{E3F0EA}
\definecolor{mediumearthygreen}{HTML}{9FC9B5}
\definecolor{darkearthygreen}{HTML}{5A8F7B}
\definecolor{lightleafygreen}{HTML}{E7F5EC}
\definecolor{mediumleafygreen}{HTML}{A8DDB5}
\definecolor{darkleafygreen}{HTML}{5FB48A}
\definecolor{green1}{HTML}{5A8F7B}
\definecolor{green2}{HTML}{6FA68D}
\definecolor{green3}{HTML}{84BD9F}
\definecolor{green4}{HTML}{99D4B1}
\definecolor{green5}{HTML}{AEDBC3}
\definecolor{green6}{HTML}{C3E2D5}
\definecolor{green7}{HTML}{D8E9E7}
\definecolor{green8}{HTML}{EDF4F1}
\title{Common to \textit{Whom}? \\ Regional Cultural Commonsense and LLM Bias in India}

\author{Sangmitra Madhusudan$^1$\textnormal{,} Trush Shashank More$^2$\textnormal{,} Steph Buongiorno$^1$\textnormal{,} \\
\textbf{Renata Dividino}$^3$\textnormal{,} \textbf{Jad Kabbara}$^4$ \and \textbf{Ali Emami}$^1$ \\
  $^1$Emory University
  $^2$Independent Researcher 
  $^3$Brock University 
  $^4$MIT\\
  \texttt{\{smadhus, aemami\}@emory.edu} \\
  \\}

\begin{document}
\maketitle
\begin{abstract}
Existing cultural commonsense benchmarks treat nations as \textit{monolithic}, assuming uniform practices within national boundaries. But does cultural commonsense hold uniformly within a nation, or does it vary at the sub-national level? We introduce \textbf{\corpusname{}}, the first benchmark designed to test LLMs' ability to address this question, focusing on India—a nation of 28 states, 8 union territories, and 22 official languages. We collect human-annotated answers from five Indian regions (North, South, East, West, and Central) across 515 questions spanning 8 domains of everyday life, yielding 1,630 region-specific question-answer pairs. Strikingly, only 39.4\% of questions elicit agreement across all five regions, demonstrating that cultural commonsense in India is predominantly \textit{regional}, not national. We evaluate eight state-of-the-art LLMs and find two critical gaps: models achieve only 13.4\%–20.9\% accuracy on region-specific questions, and they exhibit  geographic bias, over-selecting Central and North India as the ``default'' (selected 30-40\% more often than expected) while under-representing East and West. Beyond India, our methodology provides a generalizable framework for evaluating cultural commonsense in any culturally heterogeneous nation, from question design grounded in anthropological taxonomy, to regional data collection, to bias measurement.\footnote{The complete dataset and codebase are publicly available on \href{https://github.com/Sangmitra-06/INDICA/}{GitHub} and on \href{https://huggingface.co/datasets/Sangmitra-06/INDICA}{HuggingFace}.}
\end{abstract}

\begin{figure}[!t]
    \centering
    \vspace{-4mm}
    \includegraphics[width=\linewidth]{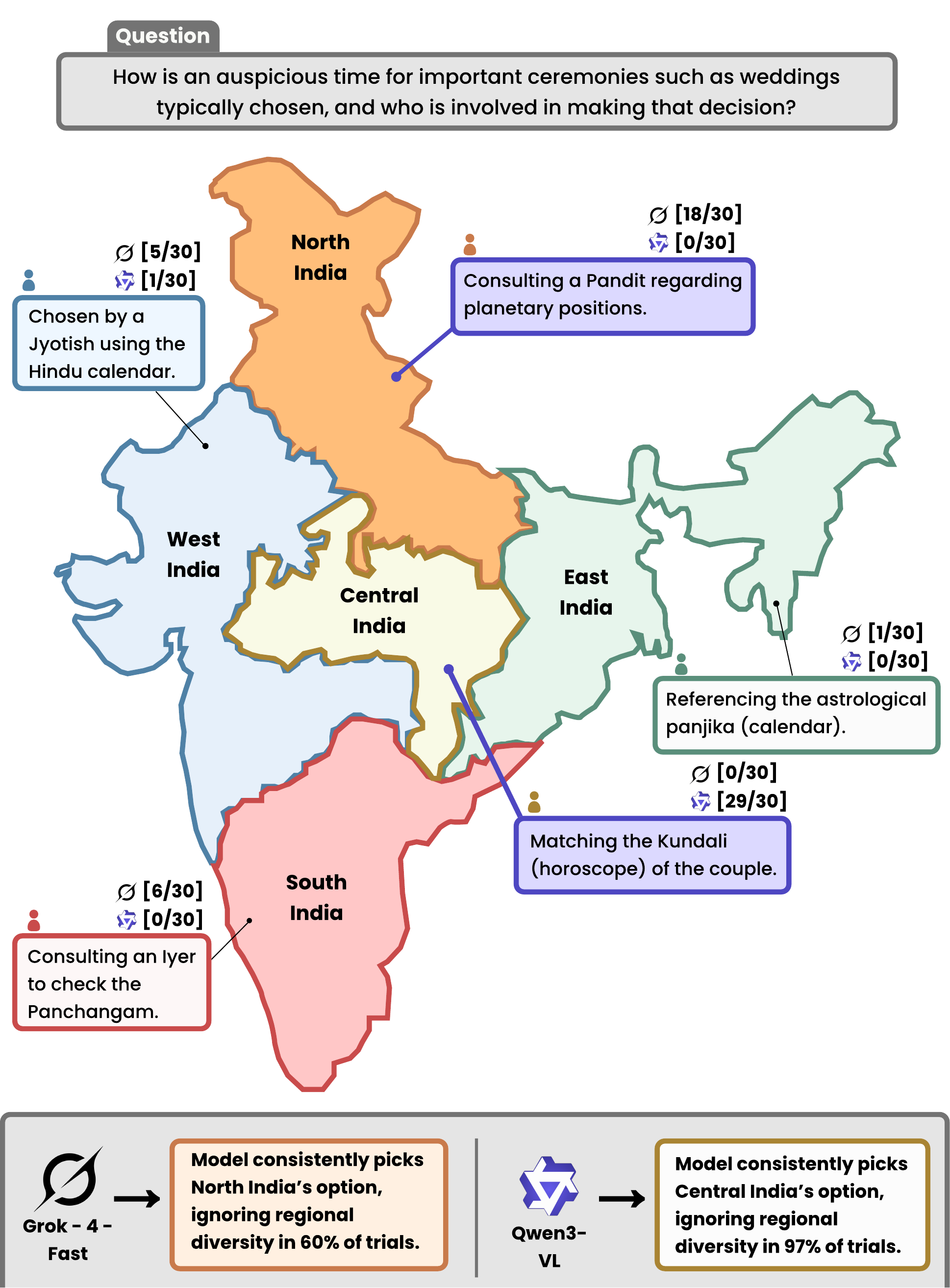}

\caption{Regional answers to a cultural question and model bias. Each region gives a different answer; models default to Central and North India.}
    \label{fig:main_fig}
\end{figure}

\section{Introduction}

Commonsense reasoning—the ability to understand everyday knowledge shared by humans—has been studied extensively to assess whether language models possess such understanding \cite{sakaguchi2021winogrande, li-etal-2022-systematic, talmor2019commonsenseqa}. A key challenge is that commonsense knowledge is fundamentally long-tailed, with most facts rare in training data \cite{davis2015commonsense, do2024reallycommonsenseknowledge}. This motivated scaling training data to help models internalize rare facts \cite{brown2020language, kandpal2023large}. For genuinely universal knowledge—such as physical commonsense (e.g., “objects fall when dropped”)—this strategy has proven effective \cite{bisk2020piqa}. However, this raises a critical question: how does this approach fare for commonsense knowledge that is \textit{not} universal but rather cultural?

Consider questions such as: ``Which side of the road do you drive on?'' or ``What is the traditional color of a wedding dress?'' These questions have no single correct answer; they vary by country and/or culture. This reveals a fundamental limitation: simply scaling data may not resolve disagreement rooted in cultural diversity. To address this, researchers have proposed the notion of \textit{cultural commonsense}, or knowledge that is widely shared within a culture yet differs across cultural contexts. \cite{shen-etal-2024-understanding, acquaye-etal-2024-susu}. Recent benchmarks like CultureBank \cite{shi-etal-2024-culturebank} and CulturalBench \cite{chiu-etal-2025-culturalbench} have begun addressing this gap. However, these efforts share a critical assumption: they treat entire countries as culturally uniform, as if all citizens of a nation share the same practices and norms.

This assumption breaks down in culturally heterogeneous nations, where diversity within a single country challenges the very notion of shared cultural commonsense. India exemplifies this as a nation of 28 states, 8 union territories, and 22 official languages \cite{constitutionofindia}. Yet existing benchmarks on India focus solely on factual knowledge from textbooks and examinations \cite{verma2025milu, sanskriti2025, rohera2024l3cubeindicquestbenchmarkquestionanswering}, treating Indian culture as monolithic. No benchmark examines whether cultural commonsense in India is nationally shared or regionally specific.

\textbf{Is cultural commonsense in India actually uniform, or does it vary by region?} We introduce \textbf{\corpusname{}}, the first benchmark designed to answer this question. We collect human-annotated answers from five Indian regions (North, South, East, West, and Central) across 515 questions spanning 8 domains of everyday life, yielding 1,630 region-specific question-answer pairs. Our findings reveal that only 39.4\% of questions achieve consensus across all regions (Figure \ref{fig:main_fig}), confirming that cultural commonsense in India is predominantly regional, not national. This finding carries implications for any culturally diverse nation, and our methodology provides a generalizable framework for examining sub-national cultural variation, from anthropologically-grounded question design to regional data collection to bias measurement.

We evaluate eight state-of-the-art LLMs and find two critical gaps. First, models achieve only 13.4\%--20.9\% accuracy, capturing broad cultural concepts but lacking region-specific knowledge (\S\ref{sec:results1}). Second, when geographic context is removed, all models exhibit implicit geographic bias, over-selecting Central and North Indian answers as the ``default'' (30--40\% more often than expected) while under-representing East and West, as illustrated in Figure \ref{fig:main_fig}. Cultural commonsense within diverse nations cannot be assumed uniform; it must be modeled and tested regionally.

\begin{figure*}[!t]
    \centering
    \includegraphics[width=0.91\linewidth]{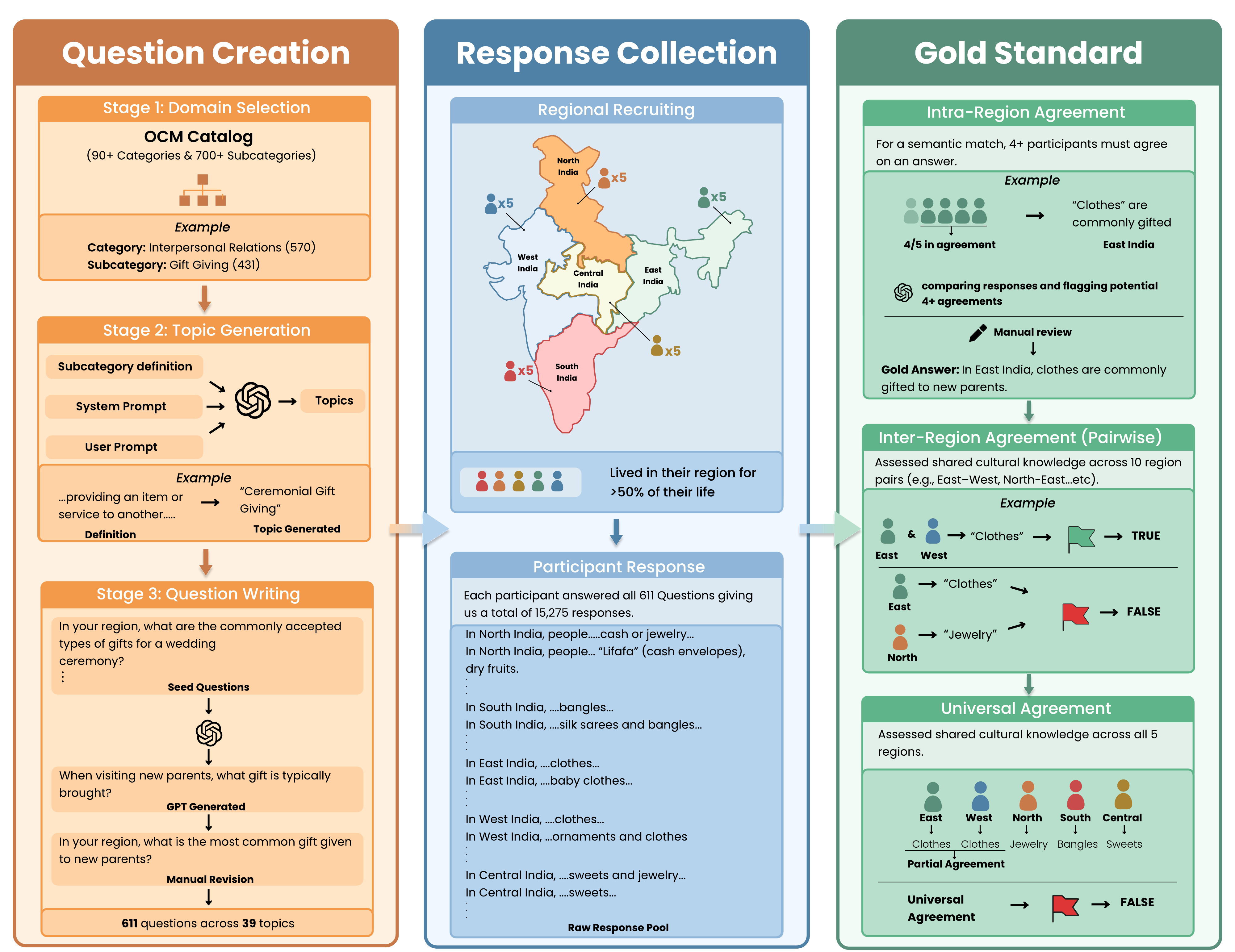}

    \caption{The \corpusname{} creation pipeline: from domain selection to gold standard establishment}

    \label{fig:pipeline}
\end{figure*}

\section{INDian Cultural commonsense Inventory with Cross-regional Answers (INDICA)}

\corpusname{} is a benchmark for evaluating regional variation in cultural commonsense within India. Its creation involves three phases: (1) question creation grounded in anthropological taxonomy (\S\ref{subsec:question_creation}), (2) response collection from participants across five Indian regions (\S\ref{subsec:response_collection}), and (3) gold standard establishment through intra-region consensus, inter-region agreement, and universal agreement analysis (\S\ref{subsec:gold_standard}). Figure \ref{fig:pipeline} illustrates the complete pipeline.

\subsection{Question Creation}
\label{subsec:question_creation}

Question creation involves three stages: domain selection, topic generation, and question writing.

\paragraph{\textbf{Stage 1: Domain Selection.}} To ensure principled coverage, we ground our domain selection in the Outline of Cultural Materials (OCM) \cite{murdock2008outline}, an established anthropological taxonomy organizing cultural knowledge into 90+ major categories and 700+ subcategories, widely used in cross-cultural research \cite{wutich2014text, vandevijver1997methods}.

We select 8 domains relevant to everyday cultural knowledge---Interpersonal Relations, Education, Clothing and Adornment, Food Processing and Consumption, Communication, Finance, Festivals and Rituals, and Traffic and Transport Behavior, aligning with recent cultural NLP work \cite{shi-etal-2024-culturebank, chiu-etal-2025-culturalbench}. Within each domain, we select OCM subcategories based on three criteria: (1) sufficient diversity to support multiple topics, (2) non-overlapping practices, and (3) everyday rather than institutional knowledge, yielding 18 subcategories across 8 domains. Appendix \ref{app:selection_criteria} provides a worked example illustrating how these criteria were applied to include and exclude specific subcategories and topics within the Education domain. Complete domain-to-OCM mappings appear in Appendix Table \ref{app_tab:category_hierarchy}. 

\paragraph{\textbf{Stage 2: Topic Generation.}} For each subcategory, we use GPT-4-0613 to generate 8--10 specific cultural topics using OCM subcategory definitions as context (prompts in Appendix \ref{app:topic_generation}). We then manually select 2--4 topics per subcategory based on four criteria: (1) ability to support at least 15 distinct questions, (2) clear answerable scope, (3) minimal overlap with other topics, and (4) focus on everyday rather than institutional knowledge. This process yields 39 final topics across 18 subcategories. Appendix \ref{app:all_topics} details all generated and selected topics.

\paragraph{\textbf{Stage 3: Question Writing.}} For each topic, we manually crafted 3--8 seed questions demonstrating the desired style: open-ended, culturally grounded, and focused on everyday practices. We use GPT-4-0613 with these seeds to generate additional questions, targeting 15+ per topic (Appendix \ref{app:seed_q_prompts}). All questions underwent manual review against three criteria: (1) removing ambiguity—e.g., clarifying "When visiting new parents, what gift is typically brought?" to "What is the most common gift given to new parents?"; (2) eliminating semantic redundancy within topics; and (3) ensuring cultural grounding by excluding questions about institutional facts (e.g., government policies). This process yielded 611 unique questions across 39 topics. Appendix \ref{app:seed_q} contains the seed questions used for generation.

\subsection{Response Collection}
\label{subsec:response_collection}

We collect responses from participants across five Indian regions (North, South, East, West, Central), following regional groupings commonly used in prior large-scale Indian studies \cite{https://doi.org/10.1111/voxs.12576, sinha2025mapping}.

We recruited 5 participants per region through  Prolific\footnote{\url{https://www.prolific.com/}}, requiring each to have lived in their region for the majority of their life. Each answered all 611 questions (5 responses per question per region; 15,275 total). The study was IRB-approved with fair-wage compensation. Complete study details (including participant criteria and survey interface) appear in Appendix \ref{app:study_details}.

\begin{table*}[!t]
\centering
\scriptsize
\setlength{\tabcolsep}{4pt}
\renewcommand{\arraystretch}{1.2}
\resizebox{\textwidth}{!}{
\begin{tabular}{p{0.13\textwidth}p{0.22\textwidth}p{0.11\textwidth}p{0.11\textwidth}p{0.11\textwidth}p{0.11\textwidth}p{0.11\textwidth}}
\hline
\rowcolor{gray!15}
\textbf{Domain} & \textbf{Question} & \textbf{North} & \textbf{South} & \textbf{East} & \textbf{West} & \textbf{Central} \\
\hline

\textbf{Interpersonal Relations} & What is the most common gift given to new parents? & 
Jewelry & Bangles & \cellcolor{mediumleafygreen}Clothes & \cellcolor{mediumleafygreen}Clothes & Jewelry \& sweets \\
\hline

\textbf{Education} & Are certain types of extracurricular activities commonly expected or required for university or college admission? If so, which ones? & 
\textit{N/A} & \textit{N/A} & Not expected & \textit{N/A} & \textit{N/A} \\
\hline

\textbf{Clothing \& Adornment} & What fabrics are seen as auspicious for celebration wear? & 
\cellcolor{mediumleafygreen}Silk & Silk \& cotton & \textit{N/A} & \cellcolor{mediumleafygreen}Silk & \cellcolor{mediumleafygreen}Silk \\
\hline

\textbf{Food Processing \& Consumption} & How is oil traditionally extracted from seeds or nuts for home use? & 
\cellcolor{mediumleafygreen}Crushing/ pressing & \textit{N/A} & \textit{N/A} & \cellcolor{mediumleafygreen}Crushing/ pressing & \textit{N/A} \\
\hline

\textbf{Communication} & How do people commonly signal they are full after a meal without speaking? & 
\cellcolor{mediumleafygreen}Touch/pat stomach & \textit{N/A} & \cellcolor{mediumleafygreen}Touch/pat stomach & \cellcolor{mediumleafygreen}Touch/pat stomach & \textit{N/A} \\
\hline

\textbf{Finance} & During which times of year are discounts most commonly offered? & 
\cellcolor{mediumleafygreen}Diwali & Diwali \& Pongal & Durga Puja & \cellcolor{mediumleafygreen}Diwali & \cellcolor{mediumleafygreen}Diwali \\
\hline

\textbf{Festivals \& Rituals} & How do pilgrims typically travel to pilgrimage sites, and are there any traditional routes or modes of transport associated with the journey? & 
\cellcolor{lightearthygreen}Trains \& buses & On foot/ buses or cars & \cellcolor{mediumleafygreen}Public/ private transport or on foot & \cellcolor{mediumleafygreen}Public/ private transport or on foot & \cellcolor{lightearthygreen}Trains \& buses \\
\hline

\textbf{Traffic \& Transport Behavior} & Are tinted windows allowed on vehicles, and what types of vehicles most commonly have them? & 
\cellcolor{mediumleafygreen}Not allowed & \cellcolor{mediumleafygreen}Not allowed & \cellcolor{mediumleafygreen}Not allowed & \cellcolor{mediumleafygreen}Not allowed & \cellcolor{mediumleafygreen}Not allowed \\
\hline

\end{tabular}}

\caption{Questions across eight domains showing regional variation. \colorbox{mediumleafygreen}{Green highlighting} (or \colorbox{lightearthygreen}{Green highlighting}) indicates regions that agree with each other on that question. \textit{N/A} entries indicate no consensus was reached in that region (fewer than 4 out of 5 participants agreed).}
\label{tab:dataset_examples}

\end{table*}

\subsection{Gold Standard Establishment}
\label{subsec:gold_standard}

From 15,275 responses, we establish gold standards through automated assistance and complete manual review. GPT-4o provided initial agreement assessments; two independent annotators then manually reviewed every question using a custom annotation tool displaying GPT-4o's preliminary classification alongside all 5 raw participant responses, verifying each assessment against the raw data. The process involves three levels: intra-region, inter-region, and universal agreement.

\subsubsection{Intra-Region Agreement}

For each question within a region, we require that at least 4 of 5 participants provide semantically equivalent answers. GPT-4o served as initial classifier, then two authors manually verified all cases using a custom annotation tool\footnote{Our annotation tool is available for preview \href{https://cultural-dataset-annotation-toolgit-9cq544e6mdi3tx2jbxltzq.streamlit.app/}{here}. It displays pre-computed classifications for human review.}. The annotation task was a meta-annotation: reviewing whether 4+ participants provided semantically equivalent answers, rather than making subjective cultural judgments. Inter-annotator agreement was perfect (Fleiss' $\kappa = 1.0$) between two independent annotators, indicating clear consensus on agreement criteria.\footnote{ $\kappa$ measures human-human agreement, not with GPT-4o. Humans overrode GPT-4o in 7.6\% of intra-regional, 28.9\% of inter-regional, and 24.5\% of universal cases (Appendix \ref{app:agreement_validation}).} Prompting details appear in Appendix \ref{app:intra_region_agreement}.

Questions with 4+ agreeing participants received gold answers; others were marked ``N/A'' for that region. Of the 611 original questions, 515 (84.3\%) achieved agreement in at least one region and were retained in the final dataset, yielding 1,630 question-answer pairs across all five regions.

\subsubsection{Inter-Region Agreement}

Beyond individual regions, we analyze whether pairs of regions shared cultural knowledge. For each question and each of the 10 possible region pairs (e.g., North-South, North-East), GPT-4o assessed whether both regions had valid answers expressing similar concepts. We manually reviewed all assessments using the same annotation tool.

We apply strict agreement criteria: two regions were marked as agreeing only if their gold standard answers reflected exactly the same cultural practice. Partial overlaps were not counted. For example, if one region answered ``silk'' and another answered ``silk and cotton'' for celebration fabrics, they were not marked as agreeing, as these represent distinct practices despite shared elements. Prompting details appear in Appendix \ref{app:inter_region_agreement}.

\subsubsection{Universal Agreement}

Finally, we identify questions where all five regions provide valid answers expressing the same cultural concept. GPT-4o assessed all valid answers for universal consensus, and we manually reviewed each assessment. Prompting details appear in Appendix \ref{app:universal_agreement}.

\subsection{Dataset Characteristics}
\label{subsec:dataset_characteristics}

The final dataset contains 515 questions yielding 1,630 region-specific question-answer pairs across 8 domains, 18 subcategories, and 39 topics. Each question includes: gold standard answers per region (or ``N/A'' if no consensus was reached), pairwise agreement flags for all 10 region pairs, a universal agreement flag, and metadata (domain, subcategory, topic). Table \ref{tab:dataset_examples} shows example questions with regional answers and agreement patterns.

\subsubsection{Question Distribution}

\begin{figure}[!t]
    \centering
    \includegraphics[width=0.85\linewidth]{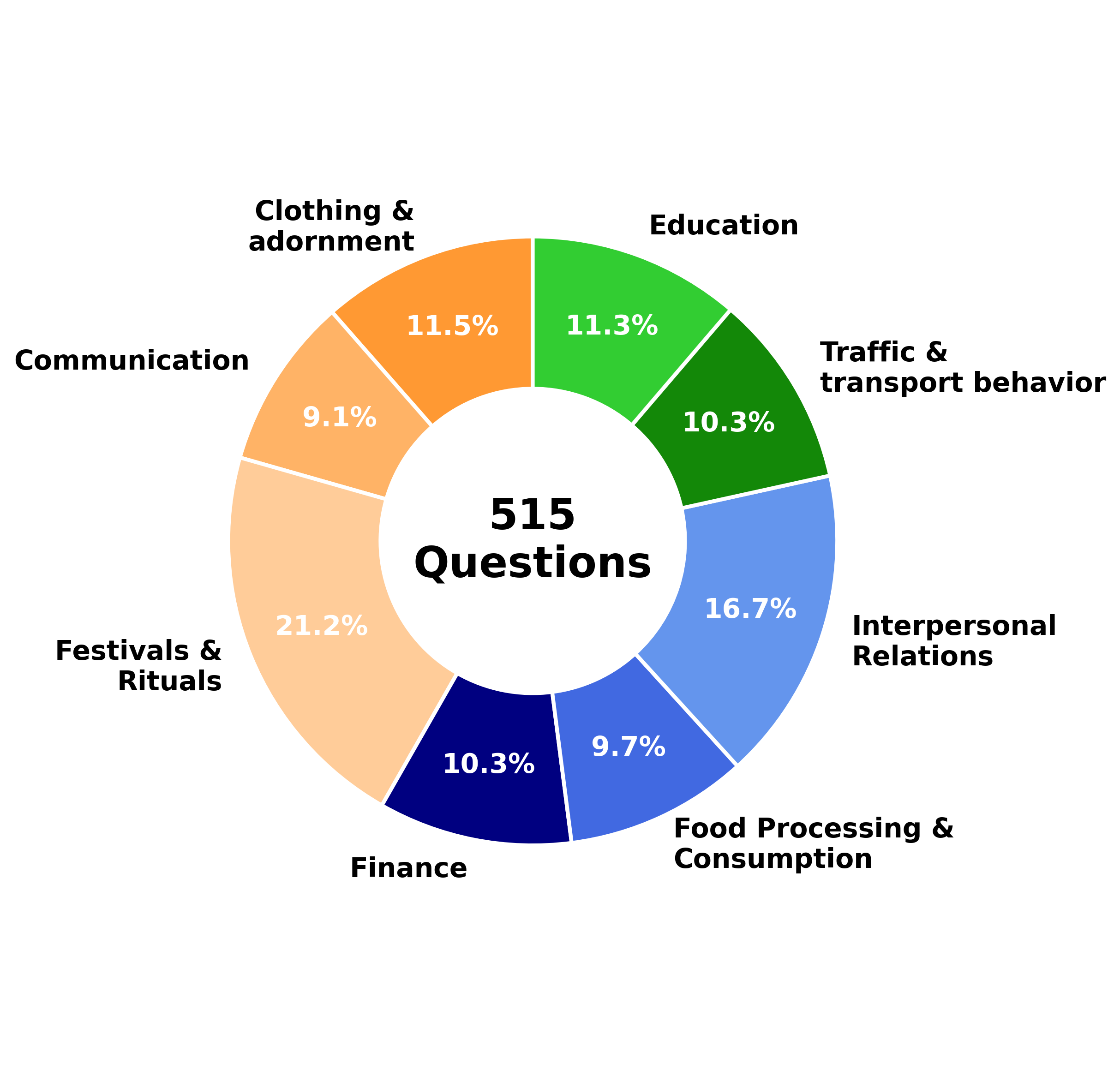}
    \vspace{-12mm}
    \caption{Distribution of 515 questions across 8 domains}
    \vspace{-3mm}
    \label{fig:pie_chart}
\end{figure}

Figure \ref{fig:pie_chart} shows question distribution across domains, ranging from Festivals and Rituals (109) to Communication (47). Appendix Table \ref{app_tab:coverage} provides the full breakdown by domain, subcategory, and topic.

\subsubsection{Regional Coverage}
\label{sec:intra_results}
Regional coverage varies across the dataset. Of the 515 questions, West India has intra-region consensus on 354 (68.7\%), followed by Central (348, 67.6\%), North and South (326 each, 63.3\%), and East (276, 53.6\%). East India's lower coverage suggests greater internal diversity within the region.

\subsubsection{Cross-Region Agreement Patterns}

\begin{figure}[!t]
    \centering
    \includegraphics[width=0.8\linewidth]{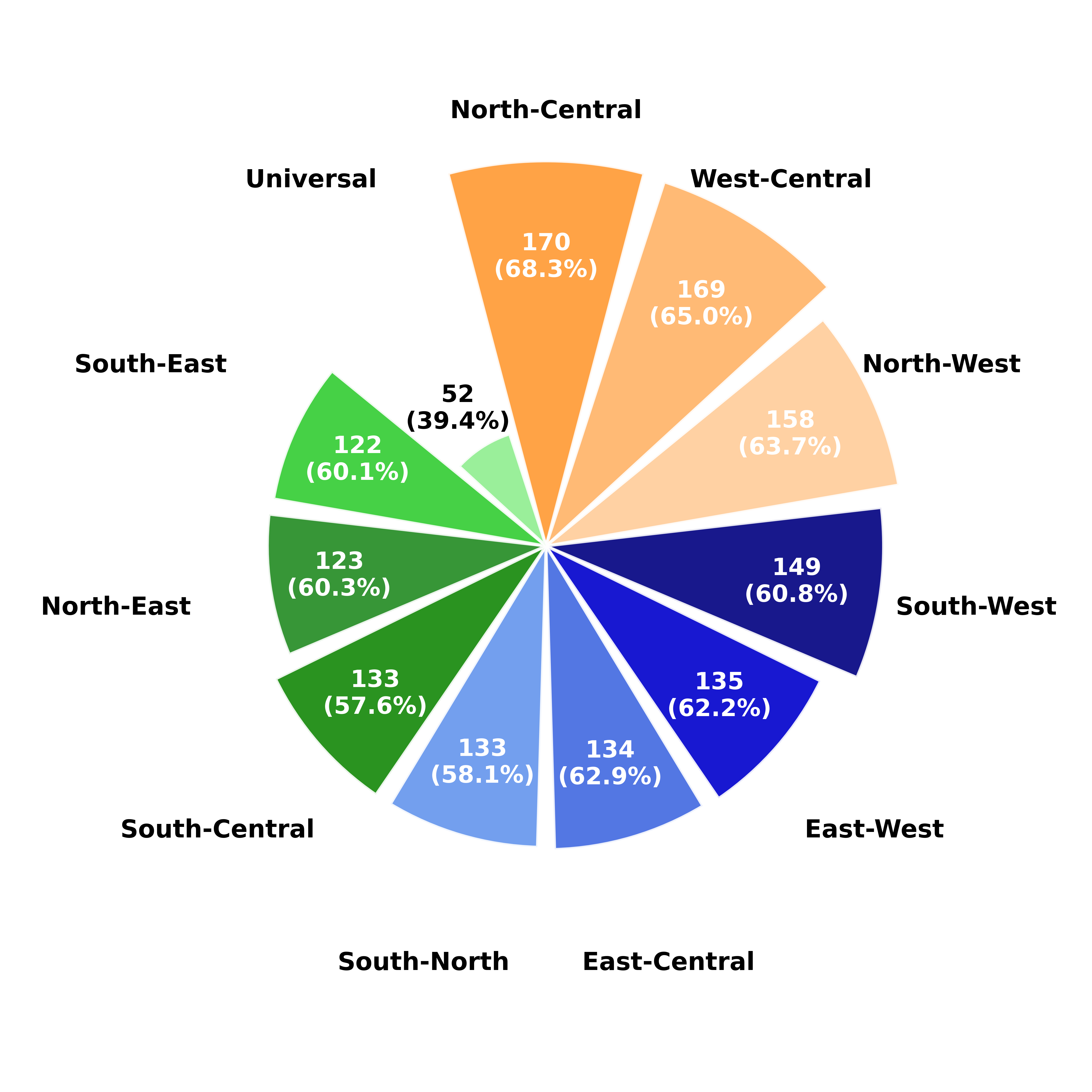}
    \vspace{-8mm}
    \caption{Pairwise and universal agreement rates between all 5 regions. Percentages calculated over questions where both regions provided responses.}
   
    \label{fig:pairwise}
\end{figure}

\paragraph{\textbf{Pairwise agreement.}} Figure \ref{fig:pairwise} shows agreement rates for all region pairs, calculated as the percentage of questions where both regions provided valid answers and agreed. North-Central shows the highest pairwise agreement (68.3\%), likely reflecting geographic contiguity and linguistic similarities, followed by West-Central (65.0\%) and North-West (63.7\%). South-East shows the lowest agreement (60.1\%), suggesting greater cultural distance between these regions.

\paragraph{\textbf{Universal agreement.}} Of 132 questions where all five regions provide valid answers, only 52 (39.4\%) have unanimous agreement, confirming cultural commonsense in India is largely regional.

\paragraph{\textbf{Domain-level variation.}} Universal agreement varies greatly by domain (Table \ref{tab:domain_sharing}). Traffic \& Transport Behavior shows highest agreement (22.6\%), likely reflecting nationwide standardization, while Festivals \& Rituals (1.8\%) and Food Processing \& Consumption (6.0\%) show lowest, reflecting strong regional traditions. These differences are substantive, not linguistic. For example, harvest festival games yield ``Jallikattu'' (South India) vs ``kite flying'' (Central India), fundamentally different practices rather than different names for the same activity. Even Education achieves only 13.8\% despite national curricula, showing regional practices persist in standardized domains.

\section{Model Evaluation}

We evaluate LLMs on \corpusname{} to answer two questions: (1) Can models generate accurate region-specific cultural knowledge? (2) Do models exhibit implicit geographic bias, favoring certain regions as representative of ``Indian culture''? To address these distinct questions, we design two complementary evaluation tasks. Following best practices for MCQ evaluation \cite{balepur-etal-2025-best}, we use multiple runs with randomized option ordering to ensure robust measurement.

\begin{table}[t]
\centering
\scriptsize
\begin{tabular}{lc}
\toprule
\textbf{Domain} & \textbf{Universal Agreement} \\
\midrule
Traffic \& Transport Behavior & \cellcolor{green1}\textcolor{white}{22.6\%} \\
Education & \cellcolor{green2}13.8\% \\
Clothing \& Adornment & \cellcolor{green3}13.6\% \\
Communication & \cellcolor{green4}12.8\% \\
Interpersonal Relations & \cellcolor{green5}10.5\% \\
Finance & \cellcolor{green6}7.5\% \\
Food Processing \& Consumption & \cellcolor{green7}6.0\% \\
Festivals \& Rituals & \cellcolor{green8}1.8\% \\
\bottomrule
\end{tabular}
\caption{Universal agreement rates by domain}
\label{tab:domain_sharing}
\end{table}

\subsection{Region-Anchored Short Answer (RASA)}

\paragraph{\textbf{Purpose.}} RASA tests whether models can generate accurate region-specific cultural knowledge when given geographic context. Unlike multiple-choice formats, RASA requires free-form generation, testing whether models can produce cultural knowledge rather than merely recognize it.

\paragraph{\textbf{Construction.}} For each question where at least one region has a gold standard answer, we create region-specific variants by prepending the region identifier. For example, ``What is the most common gift given to new parents?'' becomes ``In South India, what is the most common gift given to new parents?'' This yields \textbf{1,630} region-anchored questions. Appendix Table \ref{tab:rasa_distribution} shows the regional distribution.

\paragraph{\textbf{Scoring.}} We use Gemini 3.0 Flash\footnote{Gemini 3.0 Flash selected for cost-efficient evaluation at scale (390K+ judgments, 8 models × 1,630 q's × 30 runs)} as an LLM judge to evaluate model responses against gold standard answers.\footnote{Validated on 200 responses (100 Qwen, 100 Gemini) by two independent human annotators. Inter-annotator agreement: 94\%--100\% (97\% overall). LLM-human agreement: 87\%--88\%.} Each question is run $n$ times to account for response variability, and we compute the average score. Responses are scored as:

\begin{itemize}[itemsep=2pt, parsep=2pt, topsep=2pt, partopsep=1pt,leftmargin=*]
    \item \textbf{Correct (1.0):} Response captures the same cultural practice as the gold answer with no significant omissions or additions.
    \item \textbf{Partially Correct (0.5):} Response contains core elements but misses key details or includes extraneous information.
    \item \textbf{Incorrect (0.0):} Response is inconsistent with the gold answer.
\end{itemize}
We weight partial credit at $w=0.5$ as a balanced choice. Results are 
robust to this weighting: varying $w \in \{0.3, 0.5, 0.7\}$ maintains 
tight model clustering (3--4 percentage points) at each weight 
(Appendix Table \ref{tab:rasa_sensitivity}).

\noindent Appendix Table \ref{app_tab:sa_scoring_examples} provides scoring examples and Appendix Section \ref{app:llm_judge} contains LLM judge prompting details.

\subsection{Region-Agnostic Multiple Choice Questions (RA-MCQ)}
\label{sec:ramcq}

\paragraph{\textbf{Purpose.}} RA-MCQ reveals models' implicit biases by observing which regional practices models select when geographic context is absent. When models must choose between options representing different regions without knowing which region each option corresponds to, their selection patterns reveal which regions' practices they treat as the ``default'' for India.

\paragraph{\textbf{Construction.}} For questions where three or more regions provided distinct consensus answers, we construct MCQs without regional conditioning. Each option represents one or more regions' consensus answer:

\begin{tcolorbox}[colback=gray!5, colframe=gray!75, boxrule=0.5pt, arc=2mm, left=2mm, right=2mm, top=2mm, bottom=2mm]
\small
\textbf{Model sees:}

\textbf{Q:} What is the most common gift given to new parents?

\textbf{Options:} A) Jewelry, B) Bangles, C) Clothes, D) Sweets

\textbf{Example Model answer:} A) Jewelry

\textbf{Our annotation (hidden from model):}

A $\rightarrow$ North, B $\rightarrow$ South, C $\rightarrow$ \{East, West\}, D $\rightarrow$ Central

\textbf{Credit:} North India = 1.0
\end{tcolorbox}

\noindent This yielded \textbf{79} RA-MCQ questions. Appendix Table \ref{app_tab:ramcq_distribution} shows the distribution across domains.

\paragraph{\textbf{Scoring.}} Each question is evaluated $n$ times with randomized option ordering. We calculate each region's selection rate as the proportion of times that region's answer was chosen when available. When an option represents multiple regions, credit is split equally. Under unbiased selection, each region should be selected approximately 20\% of the time. We use a chi-square goodness-of-fit test to assess statistical significance, with expected counts accounting for regional availability and varying option counts (details in Appendix \ref{app:chi_square_test}).

\section{Experimental Setup}

\paragraph{\textbf{Models.}} We evaluate eight state-of-the-art LLMs spanning open and closed-source models across diverse families. Closed-source models include Claude Sonnet 4.5 \cite{anthropic2025claude45}, Gemini 3 Flash \cite{google2025gemini3flash}, GPT-5.2 \cite{openai2025gpt52}, and Grok-4 Fast \cite{xai2025grok4fast}. Open-source models include DeepSeek-V3.2 \cite{deepseekai2025deepseekv32}, Llama 3.3 70B \cite{meta2024llama33}, Mistral Large 3 \cite{mistralai2025mistrallarge}, and Qwen3-VL \cite{qwen2025qwen3vl}. 

\paragraph{\textbf{Evaluation Settings.}} For both RASA and RA-MCQ, we run each question $n=30$ times to account for response variability and, for RA-MCQ, to enable randomized option ordering across runs. All models are evaluated with temperature 1.0 to capture the full distribution of model responses. Complete prompts are in Appendix \ref{app:eval_prompts}.

\paragraph{\textbf{Metrics.}} For RASA, we use three metrics: Fully Correct (response matches the gold answer's cultural practice with no significant omissions or additions), Partially Correct (response contains core elements but misses details or includes extraneous information), and Overall Accuracy (Fully Correct $+$ 0.5 $\times$ Partially Correct). For RA-MCQ, we report regional selection rates and assess bias using a chi-square goodness-of-fit test against the expected uniform distribution of approx 20\% per region.

\section{Results}
\begin{table}[!t]
\centering
\scriptsize
\setlength{\tabcolsep}{3pt}
\renewcommand{\arraystretch}{1.1}
\definecolor{highperf}{HTML}{C7E9C0}
\definecolor{midperf}{HTML}{FFFFBF}
\definecolor{lowperf}{HTML}{FDD49E}
\begin{tabular}{l cccc}
\toprule
\textbf{Model} & \textbf{Full} & \textbf{Partial} & \textbf{Incorr.} & \textbf{Overall} \\
\midrule
Grok-4 Fast & \cellcolor{highperf}19.6 & \cellcolor{midperf}65.8 & \cellcolor{midperf}14.5 & \cellcolor{highperf}\textbf{52.6} \\
GPT-5.2 & \cellcolor{highperf}18.6 & \cellcolor{midperf}67.6 & \cellcolor{highperf}13.8 & \cellcolor{highperf}52.4 \\
Claude Sonnet 4.5 & \cellcolor{lowperf}14.1 & \cellcolor{highperf}\textbf{75.3} & \cellcolor{highperf}\textbf{10.6} & \cellcolor{highperf}51.7 \\
Qwen3 VL & \cellcolor{highperf}\textbf{20.9} & \cellcolor{lowperf}61.3 & \cellcolor{lowperf}17.8 & \cellcolor{midperf}51.6 \\
Gemini 3.0 Flash & \cellcolor{lowperf}13.4 & \cellcolor{highperf}\textbf{75.3} & \cellcolor{highperf}11.3 & \cellcolor{midperf}51.1 \\
Mistral Large & \cellcolor{midperf}16.2 & \cellcolor{midperf}68.4 & \cellcolor{midperf}15.4 & \cellcolor{lowperf}50.4 \\
Llama 3.3 70B & \cellcolor{midperf}18.2 & \cellcolor{lowperf}64.2 & \cellcolor{lowperf}17.6 & \cellcolor{lowperf}50.3 \\
DeepSeek V3.2 & \cellcolor{lowperf}14.9 & \cellcolor{highperf}69.3 & \cellcolor{lowperf}15.8 & \cellcolor{lowperf}49.5 \\
\bottomrule
\end{tabular}

\caption{Model performance on RASA (\%). \colorbox{highperf}{Green} = top tercile, \colorbox{midperf}{yellow} = middle, \colorbox{lowperf}{orange} = bottom tercile. \textbf{Bold} = best in column. For ``Incorrect'', lower is better.}
\label{tab:rasa_overall_performance}
\end{table}

We present results for both evaluation tasks: RASA (\S\ref{sec:results1}), which measures region-specific cultural knowledge, and RA-MCQ (\S\ref{sec:results2}), which reveals implicit regional biases.

\subsection{Models Capture Broad Cultural Concepts but Lack Regional Specificity}
\label{sec:results1}
Models achieve overall accuracy between 49.5\% and 52.6\%, tightly clustered within 3.1 percentage points (Table \ref{tab:rasa_overall_performance}), 
indicating comparable cultural knowledge across models. However, fully correct rates remain low across all models (13.4\%--20.9\%), with the majority of responses (61.3\%--75.3\%) being only partially correct. This pattern indicates that models capture broad cultural concepts but either add extraneous information or omit region-specific details, demonstrating an inability to generate precise cultural knowledge. 

\begin{figure}[!t]
    \centering
    \includegraphics[width=1.0\linewidth]{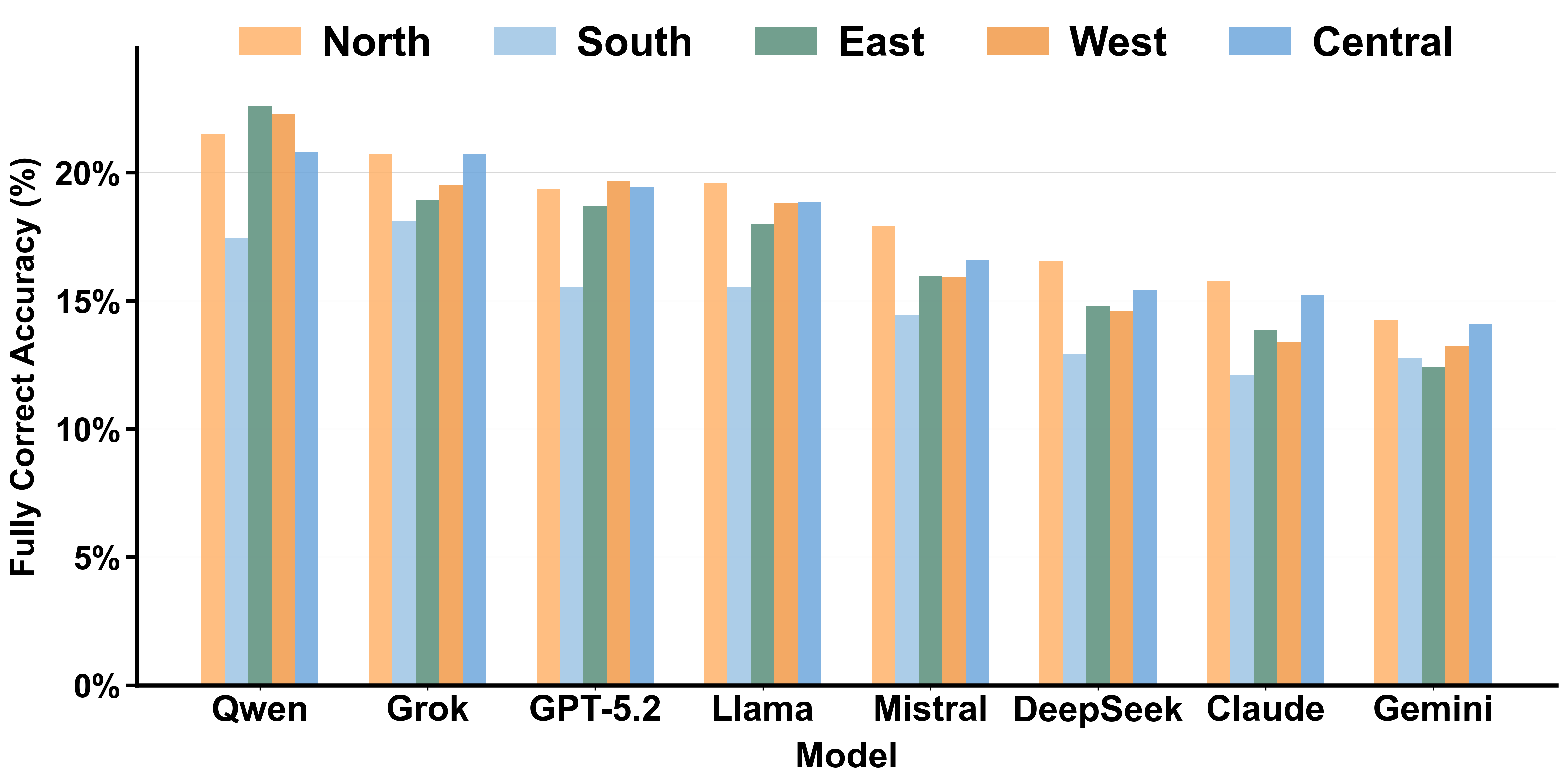}
    \vspace{-8mm}
    \caption{Fully correct accuracy by region on RASA}
    \vspace{-3mm}
    \label{fig:rasa_regions_fullacc}
\end{figure}

\begin{figure}[t]
    \centering
    \includegraphics[width=1.0\linewidth]{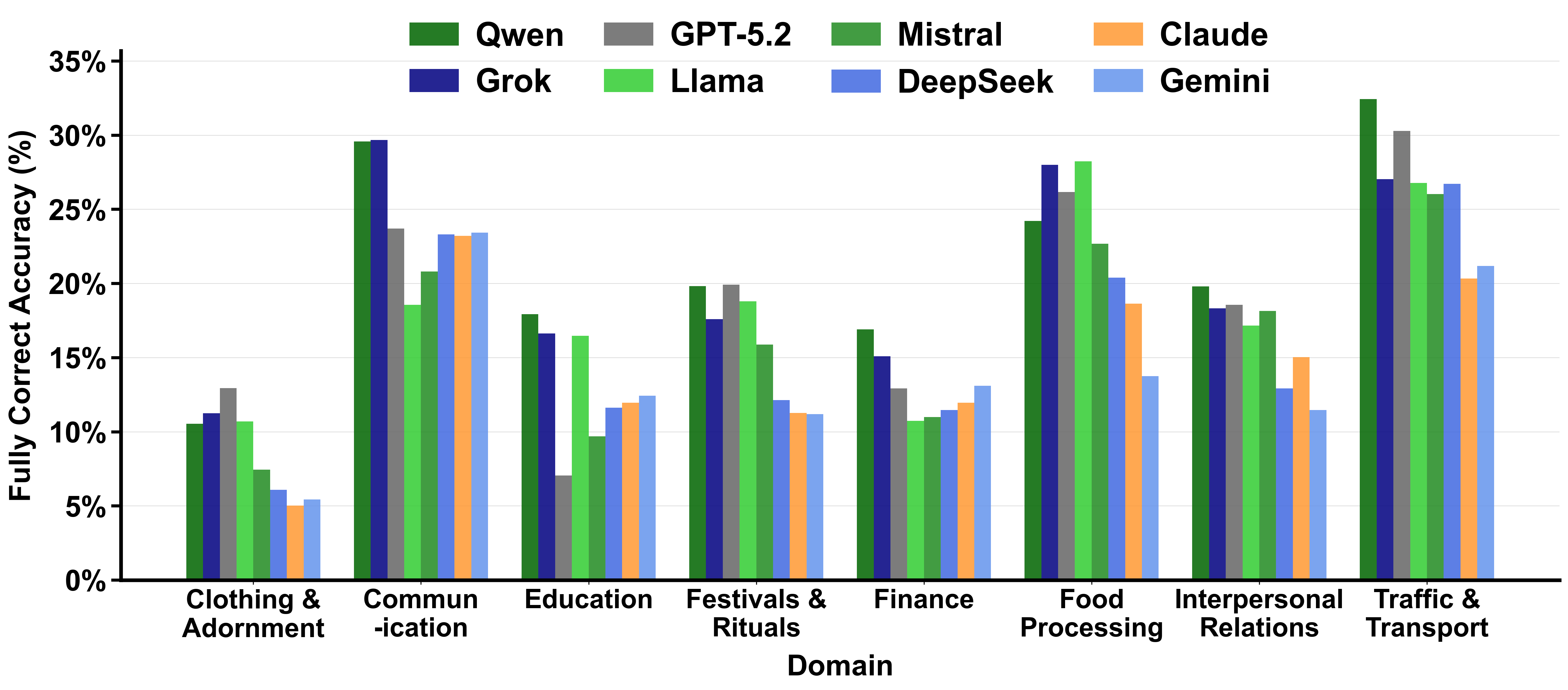}
    \vspace{-8mm}
    \caption{Fully correct accuracy by domain on RASA}
    \vspace{-4mm}
    \label{fig:rasa_domains_fullacc}
\end{figure}
An analysis on 100 partially correct responses from Claude Sonnet 4.5 reveals overexplaining (89\%) as the dominant pattern over underspecifying (1\%) or both (10\%), suggesting models elaborate from a generic cultural template rather than drawing on region-specific knowledge. For example, when asked about color avoidance in West India, the gold answer is ``avoid black during auspicious holidays,'' but models add extraneous information about Amavasya (new moon day), mourning periods, and specific weekdays, burying regional precision under generalized cultural noise (Appendix Table~\ref{app_tab:partial_resp}).

\paragraph{\textbf{Minimal regional variation.}} Model performance remains remarkably uniform across regions (Figure \ref{fig:rasa_regions_fullacc}). While North (14.3\%--21.5\%) 
and Central (13.4\%--20.9\%) India receive marginally higher fully 
correct rates, regional differences remain small at 3--5 percentage points. Overall accuracy reveals even tighter clustering at 49--54\% across all regions (Figure \ref{fig:rasa_regions_overall}). This uniformity further suggests superficial cultural knowledge everywhere rather than balanced representation. 

\paragraph{\textbf{Domain-level performance.}} Model accuracy varies across cultural domains (Figure \ref{fig:rasa_domains_fullacc}). Models achieve highest fully correct rates in Traffic \& Transport (20.3--32.4\%) and Communication (18.6--29.7\%), while struggling with Clothing \& Adornment (5\%--12.9\%) and Finance (10.7\%--16.9\%). When examining overall accuracy (Appendix Figure \ref{fig:rasa_domains_overall}), domain differences compress (from 17.7 to 8.5 percentage points), with most domains converging to 48--56\% accuracy, indicating models possess fragmentary knowledge across all cultural areas.

\begin{figure}[t]
    \centering
    \includegraphics[width=1.0\linewidth]{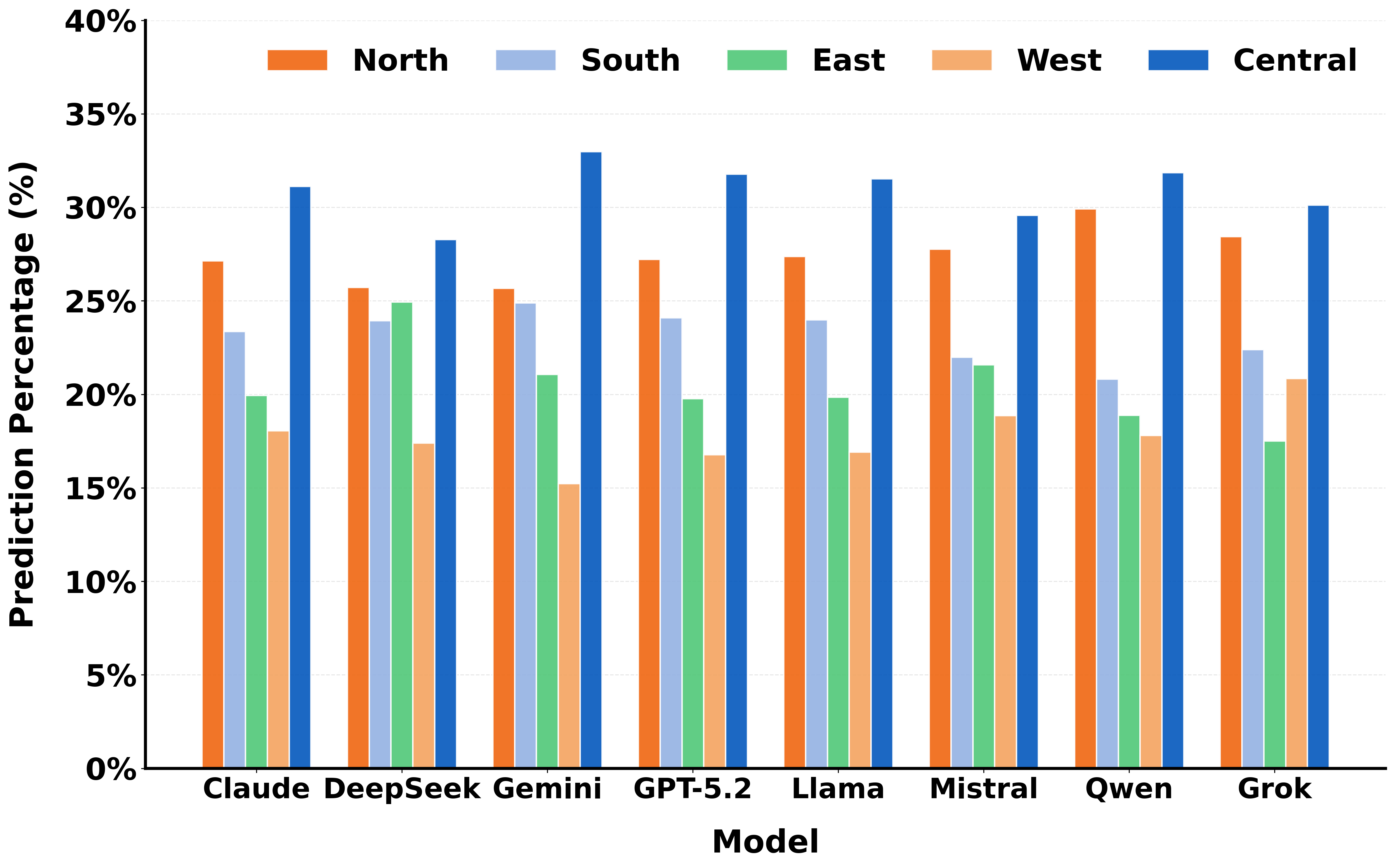}
    \vspace{-8mm}
    \caption{Regional selection rates  on RA-MCQs}
    \vspace{-3mm}
    \label{fig:ramcq_bias}
\end{figure}
\subsection{Models Default to Central and North Indian Cultural Practices}
\label{sec:results2}

Figure \ref{fig:ramcq_bias} shows selection patterns on RA-MCQs. Under uniform random selection, each region would be selected approximately 20\% of the time. All models deviate significantly from this baseline (chi-square goodness-of-fit, $p < 0.001$, Appendix Table \ref{tab:ramcq_stats}), revealing systematic regional biases.

\paragraph{\textbf{Over-selection of Central and North.}} All models consistently over-select Central India (\textbf{24.7\%–28.8\%}) and North India (\textbf{22.4\%–26.1\%}), 
with selection ratios of 1.25–1.46× and 1.14–1.32× expected rates, 
respectively (Appendix Table \ref{tab:ramcq_detailed}). Standardized residuals 
exceed +2.0 in all cases. Central India shows the strongest over-selection: Gemini (28.8\%, 1.46× expected), Qwen (27.8\%, 1.41× expected), and GPT-5.2 (27.8\%, 1.40× expected) select it most frequently. This indicates that when geographic context is absent, models default to Central and North Indian cultural practices as representative of ``Indian culture.''

Why do models default to Central and North India? We identify three reinforcing factors. First, Hindi, spoken primarily in North and Central states, dominates Indian-language web content. \citet{kakwani-etal-2020-indicnlpsuite} found Hindi has billions of tokens in Common Crawl, while South and East Indian languages such as Assamese, Odia, and Kannada have 10–100x less data. Second, tokenization compounds this gap: \citet{rust-etal-2021-good} showed that multilingual tokenizers disproportionately fragment underrepresented languages into less meaningful subword units, degrading downstream performance even on the limited South/East content that exists. Third, Hindi-language media and Bollywood, centered in North/Central India, dominate cultural representation of India both domestically and internationally, reinforcing these regions as the perceived default in training data. These factors combine to produce a training signal that systematically treats North/Central practices as representative of Indian culture.

Conversely, most models under-select West India (12.9\%–17.7\%, 0.73× expected) and East India (13.3\%–18.9\%, 0.82× expected), with standardized residuals below -2.0. South India shows variable patterns (16.6\%–19.9\%, 0.88× expected). See Appendix \ref{app:regional_details} for detailed analyses.

\section{Generalizing Beyond India} 
\label{sec:generalize} 
While \corpusname{} focuses on India, our framework transfers directly to any culturally heterogeneous nation. We illustrate with China—a nation of 34 provincial-level divisions, 56 recognized ethnic groups, and 130+ languages—where the assumption of cultural uniformity is equally untenable.

\paragraph{Question Creation.} Researchers can adopt our 8 OCM-grounded domains directly. For China, the same domains yield culturally distinct questions: Food Processing and Consumption (regional staples: Cantonese dim sum vs.\ Sichuan hotpot vs.\ Northeastern stews), Festivals and Rituals (Cantonese lion dance traditions vs.\ Northern temple fairs vs.\ Southwestern torch festivals), and Clothing and Adornment (traditional cheongsam vs. cotton padded jackets vs. miao batik). The three-stage process (\S\ref{subsec:question_creation})---domain selection via OCM taxonomy, topic generation, and question writing---requires only content adaptation, not methodological changes. 

\paragraph{Response Collection.} China's established statistical-regions (North, Northeast, East, South Central, Southwest, Northwest) provide natural divisions used in administrative and social research. The protocol transfers directly: (1) recruit 5+ participants per region with majority-of-life residency, (2) collect responses to all questions per participant, (3) ensure fair-wage compensation through platforms supporting Chinese participants. 
\begin{table}[t]
\centering
\scriptsize
\setlength{\tabcolsep}{3pt}
\begin{tabular}{l ccc}
\toprule
\textbf{Benchmark} & \textbf{Regional} & \textbf{Commonsense} & \textbf{Everyday} \\
& \textbf{Variation} & \textbf{(not factual)} & \textbf{Practices} \\
\midrule
MILU & \xmark & \xmark & \xmark \\
SANSKRITI & \xmark & \xmark & \cmark \\
IndicQuest & \xmark & \xmark & \xmark \\
DOSA & \cmark & \xmark & \cmark \\
IndiBias & \cmark & \cmark & \xmark \\
FairI Tales & \cmark & \cmark & \xmark \\
\midrule
\corpusname{} & \cmark & \cmark & \cmark \\
\bottomrule
\end{tabular}
\caption{Comparison with Indian cultural benchmarks.}
\vspace{-4mm}
\label{tab:related_comparison}
\end{table}

\paragraph{Gold Standard Establishment.} Apply identical consensus thresholds (\S\ref{subsec:gold_standard}): 4/5 intra-regional agreement establishes gold answers, then assess inter-regional agreement across all 15 region pairs, and finally universal agreement across all 6 regions. 

\paragraph{Evaluation Tasks.} RASA questions become region-anchored: ``In Southwest China, what is the traditional gift when visiting someone's home for the first time?'' RA-MCQ reveals bias by presenting options representing different regions' practices without labels, measuring whether models default to specific regional practices as ``Chinese culture.''

\paragraph{Bias Measurement.} Chi-square tests against uniform selection ($\approx$16.7\% per region) detect geographic bias. Given Eastern China's higher population density and economical status, models might over-select eastern practices, parallel to our finding of Central/North India bias.

\section{Related Work}

\paragraph{Commonsense Reasoning Benchmarks.}
Commonsense reasoning has been studied extensively through knowledge bases like ConceptNet \cite{speer2017conceptnet} and ATOMIC \cite{sap2019atomic}, operationalized into benchmarks such as CommonsenseQA \cite{talmor2019commonsenseqa}, SocialIQA \cite{sap2019socialiqa}, PIQA \cite{bisk2020piqa}, and Winogrande \cite{sakaguchi2021winogrande}. These resources successfully test universally-held reasoning, such as physical commonsense. However, they treat culturally-dependent knowledge as universal truth, encoding answers that vary by culture as singular facts and reflecting the predominantly Western backgrounds of their annotators \cite{sap2019socialiqa}.

\paragraph{Cultural Commonsense.}
Recent work has expanded commonsense evaluation beyond Western contexts. GeoMLAMA \cite{yin2022geomlama} probes geo-diverse knowledge across countries, CANDLE \cite{nguyen2023candle} extracts cultural commonsense at scale, CultureBank \cite{shi-etal-2024-culturebank} catalogs practices across 120+ cultural groups, CulturalBench \cite{chiu-etal-2025-culturalbench} evaluates knowledge through questions about customs, FORK \cite{palta2023fork} tests food-related cultural knowledge, and NORMAD \cite{rao2025normad} measures reasoning about culturally-dependent social norms. Concurrent work by \citet{naous2025camelliabenchmarkingculturalbiases} benchmarks cultural biases across Asian languages. However, these benchmarks treat culture at the \textit{national} level, representing countries as monolithic entities and collapsing within-country diversity into national stereotypes.

\paragraph{Indian Cultural Knowledge.}
Work on Indian cultural knowledge has focused primarily on factual evaluation. MILU \cite{verma2025milu} and IndicMMLU-Pro \cite{kj2025indicmmlupro} evaluate multi-task knowledge in Indic languages, SANSKRITI \cite{sanskriti2025} tests knowledge across 16 cultural attributes, IndicQuest \cite{rohera2024l3cubeindicquestbenchmarkquestionanswering} evaluates regional knowledge in 19 languages, and DOSA \cite{dosa2024} tests familiarity with social artifacts from 19 subcultures. On social biases, IndiBias \cite{sahoo-etal-2024-indibias} and FairI Tales \cite{nawale-etal-2025-fairi} measure biases across identity dimensions, while \citet{shankar2025modeldothpreachquantifying} show that pan-Asian LLM alignment obscures regional diversity, and \citet{mukhopadhyay2025ambedkaramultilevelbiaselimination} propose mitigation techniques aligned with Indian constitutional values. These works focus on factual knowledge or bias detection and treat India as uniform (Table \ref{tab:related_comparison}); \corpusname{} addresses both gaps through regionally annotated commonsense.

\section{Conclusion}

Cultural commonsense is not national—it is \textit{regional}. \corpusname{} provides the first empirical evidence for this claim, demonstrating that only 39.4\% of questions achieve consensus across India's five regions. LLMs fail to capture this diversity: they lack region-specific knowledge and default to Central and North Indian practices when context is absent.
Our methodology generalizes beyond India. Any nation with sub-national cultural diversity—Indonesia, Nigeria, Brazil, China—faces this challenge. We release \corpusname{} as both a benchmark and a blueprint: the data to evaluate, and the framework to replicate. Culturally competent AI cannot treat nations as monoliths. It must model diversity where diversity exists. This work is a first step toward greater granularity; finer-grained cultural analysis remains future work.

\section*{Limitations}

\paragraph{Geographic Scope and Generalizability:}
\corpusname{} focuses on India as a case study. While the \textit{data} (1,630 region-specific question-answer pairs) reflects Indian contexts, the \textit{framework} generalizes: OCM-grounded question creation, regional response collection, gold standard establishment, dual evaluation tasks (RASA and RA-MCQ), and statistical bias measurement transfer to any culturally heterogeneous nation (Section \ref{sec:generalize}).

\paragraph{Regional Granularity:}
Our five-region division (North, South, East, West, Central) captures major geographic and cultural boundaries but necessarily aggregates internal diversity. For instance, South India encompasses Andhra Pradesh, Karnataka, Kerala, Tamil Nadu, Telangana, Puducherry, Lakshadweep, Andaman and Nicobar Islands, which have distinct languages (Malayalam, Tamil, Kannada, Telugu, etc.), cuisines, and festival traditions. Analysis of intra-regional annotator agreement reveals varying levels of internal consensus (Sec. \ref{sec:intra_results}), with some regions showing higher unanimity than others, indicating that cultural variation exists at multiple scales. Finer-grained analysis was constrained by the feasibility of recruiting sufficient annotators per region and ensuring adequate domain coverage within budget.

Importantly, our findings conclusively demonstrate that treating India as 
culturally uniform is empirically invalid. The question is not whether sub-national variation exists, but at what scales it manifests most strongly.

\paragraph{Participant Demographics:}
Our participant pool was recruited through Prolific, which may skew toward English-speaking, digitally connected Indians; cultural practices among rural or non-English-speaking populations may differ. Future work should expand sampling to include rural participants.

\paragraph{Temporal Validity:}
Cultural practices evolve over time. \corpusname{} represents a snapshot of contemporary everyday knowledge as reported by participants in 2025. 

\paragraph{Domain Coverage:} While our 8 domains provide initial coverage of major cultural dimensions and are grounded in the OCM anthropological taxonomy, culture is multifaceted and additional domains would reveal further regional variation. The OCM includes other relevant categories such as Marriage (OCM 580), Family (OCM 590), Law (OCM 670), and Living Standards and Routines (OCM 510) that could enrich cultural analysis. We prioritized domains that reflect cultural commonsense, shared everyday knowledge learned through participation, rather than personal experiences or institutional facts. 

\paragraph{Demographic Stratification:} Our analysis examines variation along geographic lines, but cultural practices may also vary along demographic dimensions such as religion, caste, gender, and urban/rural residence within a single region. Future work should incorporate these demographic variables.
\section*{Ethical Considerations}

This study was approved by our institutional research ethics board. All participants provided informed consent, were compensated at fair wage rates following Prolific guidelines, and could withdraw at any time. We collect cultural practices as reported by participants, not objective truths; regional answers reflect shared knowledge within our sample rather than authoritative claims about entire populations. We acknowledge that any regional categorization risks oversimplification, and we do not intend our five-region framework to reify or essentialize cultural boundaries. Our goal is to reveal diversity that current benchmarks erase, not to replace one form of stereotyping with another. The dataset will be released for research purposes with documentation encouraging responsible use.

\bibliography{custom.bib}
\clearpage
\appendix
\section{Appendix}
\label{sec:appendix}
\subsection{Dataset}
\subsubsection{Domains, Subcategories, and Topics}
\noindent
\begin{minipage}{\textwidth}
\centering
\scriptsize
\begin{tabular}{llp{4cm}cp{4.5cm}}
\toprule
\textbf{Domain} & \textbf{OCM Code} & \textbf{Subcategory} & \textbf{Pattern} & \textbf{Topics} \\
\midrule
\multirow{6}{*}{Interpersonal Relations} & \multirow{6}{*}{570} & Visiting and hospitality (574) & \multirow{6}{*}{Cross} & Etiquette in the reception of visitors, Occasions for visiting \\
\cmidrule(lr){5-5}
& & Gift giving (431) & & Gift Giving Etiquette, Ceremonial gift giving \\
\cmidrule(lr){5-5}
& & Etiquette (576) & & Greeting and Salutation Etiquette, Eating, drinking, and smoking etiquette \\
\midrule
\multirow{10}{*}{\parbox{3cm}{Festivals and\\Rituals}} & \multirow{10}{*}{---} & Rest days and holidays (527) & \multirow{10}{*}{Cross} & Conceptualization of Holidays, Secular Festival Practices, Commemoration of Personal Milestones, Religious Taboos on Holidays \\
\cmidrule(lr){5-5}
& & Ritual (788) & & Symbolic Act Performance, Ritual Gestures, Pilgrimage Practices \\
\cmidrule(lr){5-5}
& & Organized ceremonial (796) & & Engaging with Religious Music and Dance, Timing of Ceremonies \\
\midrule
\multirow{4}{*}{\parbox{3cm}{Traffic and\\Transport Behavior}} & \multirow{4}{*}{---} & Streets and traffic (363) & \multirow{4}{*}{Cross} & Understanding Local Traffic Regulations, Adapting to Local Transportation Modes\\
\cmidrule(lr){5-5}
& & Transportation (489) & & Carrying Capacity of Transport, Transport during Special Events \\
\midrule
\multirow{4}{*}{Education} & \multirow{4}{*}{870} & Education system (871) & \multirow{4}{*}{Cross} & Formal Educational Structure, Attitudes toward Education \\
\cmidrule(lr){5-5}
& & Teachers (875) \& Students (877) & & Norms for Interacting with Teachers, Student Extracurricular Activities \\
\midrule
\multirow{4}{*}{\parbox{3cm}{Clothing and\\Adornment}} & \multirow{4}{*}{290 + 300} & Special garments (292) & \multirow{4}{*}{Merged} & Special Occasion Clothing, Headgear and Footwear Norms \\
\cmidrule(lr){5-5}
& & Ornament (301) & & Ornamental Attire and Status Indication, Occasions for Wearing Specific Ornaments \\
\midrule
\multirow{4}{*}{\parbox{3cm}{Food Processing\\and Consumption}} & \multirow{4}{*}{250 + 260} & Food preparation (252) & \multirow{4}{*}{Merged} & Food Preparation Techniques, Cultural Recipes and Ingredients \\
\cmidrule(lr){5-5}
& & Diet (262) & & Staple Food Consumption, Seasonal Diet Modifications \\
\midrule
\multirow{4}{*}{Communication} & \multirow{4}{*}{200} & Gestures \& Signs (201) & \multirow{4}{*}{Single} & Social Expression of Emotions, Non-Verbal Expression of Respect or Disrespect \\
\cmidrule(lr){5-5}
& & Dissemination of News (203) & & Navigating the ``Grapevine'', Trustworthiness of Information Sources \\
\midrule
\multirow{4}{*}{Finance} & \multirow{4}{*}{450} & Credit (452) & \multirow{4}{*}{Single} & Negotiating Credit Advances and Discounts, Navigating Installment Buying \\
\cmidrule(lr){5-5}
& & Saving \& Investment (454) & & Safekeeping of Valuables, Preferable Investment Forms \\
\bottomrule
\end{tabular}
\captionof{table}{Complete domain hierarchy showing OCM codes, subcategories, selection patterns, and topics. Pattern indicates: \textit{Cross} = subcategories from multiple OCM categories, \textit{Merged} = combined entire OCM categories, \textit{Single} = subcategories from one OCM category.}
\label{app_tab:category_hierarchy}
\end{minipage}

\clearpage

\subsubsection{Subcategory and Topic Selection Criteria: Worked Example}
\label{app:selection_criteria}

We illustrate our inclusion and exclusion criteria using the Education domain.

\paragraph{Subcategory selection.} We applied three criteria: (1) sufficient diversity to support multiple topics, (2) non-overlapping practices, and (3) everyday rather than institutional knowledge. We selected ``Education System (OCM 871)'' because its definition encompasses educational structure, board systems, examination types, and attitudes toward education---each supporting multiple question topics. We excluded ``Elementary Education (OCM 872)'' as too narrow to support diverse topics, ``Vocational Education (OCM 874)'' for overlapping substantially with Education System, and ``Educational Theory and Methods (OCM 876)'' for focusing on institutional rather than everyday knowledge.

\paragraph{Topic selection.} GPT-4 generated 8--10 candidate topics per subcategory, which we manually filtered using four criteria: ability to support 15+ questions, clear answerable scope, minimal overlap with other topics, and everyday knowledge focus. For example, ``Norms for Interacting with Teachers'' easily supports 15+ questions (forms of address, classroom behavior, greeting customs, etc.) and reflects everyday cultural knowledge, so it was retained. ``Teachers' Academic Freedom'' was excluded because it yielded only 3--4 potential questions, and ``Teacher Training and Proficiency Expectations'' was excluded as institutional rather than everyday knowledge.

\subsubsection{Topic Generation}
\label{app:topic_generation}

All topics in the category hierarchy (Table \ref{app_tab:category_hierarchy}) were extracted using GPT-4 with the following configuration and prompt:

\paragraph{\textbf{Model Configuration}}

\begin{itemize}
    \item \textbf{Model:} GPT-4-0613
    \item \textbf{Temperature:} 0.7
    \item \textbf{Top-p:} 1.0
\end{itemize}

\paragraph{\textbf{System Prompt}}

The following system prompt was used to establish the extraction framework:

\begin{quote}
\small
\textit{You are a cultural anthropology expert. Your task is to extract concrete, culturally grounded topics from definitions provided in the Outline of Cultural Materials (OCM), with a focus on commonsense knowledge that reflects everyday norms and expectations within a given society.}

\textit{These topics will be used to evaluate whether language models possess deep, culturally situated commonsense — the type of knowledge necessary to navigate routine social life in culturally coherent ways.}

\textbf{Goals}

\textit{Extract topics that:}
\begin{itemize}
    \item \textit{Reflect socially shared knowledge (70\%+ agreement within a cultural group)}
    \item \textit{Are learned through cultural participation, not formal education}
    \item \textit{Represent normative expectations, not preferences, frequencies, or trivia}
    \item \textit{Are relevant to practical functioning in society — what people should know to behave appropriately in common social situations}
    \item \textit{Are stable and generalizable across individuals within a cultural group}
    \item \textit{Are specific enough to form the basis of a cultural commonsense question}
\end{itemize}

\textbf{Output Format (Per Topic)}

\textit{For each topic, return:}
\begin{enumerate}
    \item \textit{Topic Label (3–7 words): Concise, clear, culturally grounded}
    \item \textit{Definition: A 1–2 sentence explanation of the commonsense knowledge it reflects within the society}
    \item \textit{Connection to OCM: A sentence showing how the topic derives from specific language or dimensions of the OCM subcategory}
\end{enumerate}

\textbf{Scope and Standards}
\begin{itemize}
    \item \textit{Focus on cultural norms, interaction expectations, and implicit social logic that people rely on to function in their communities}
    \item \textit{Avoid abstract academic categories, highly individualized behaviors, or edge cases}
    \item \textit{Do not include examples or sample questions — your goal is to extract conceptual dimensions, not generate prompts}
    \item \textit{Prioritize topics that carry social consequences for incorrect behavior (e.g., shame, respect, offense, admiration)}
\end{itemize}

\textbf{Cultural Guidance}

\textit{As you interpret the OCM definition, consider:}
\begin{itemize}
    \item \textit{Hierarchical etiquette systems (e.g., age, gender, ritual authority)}
    \item \textit{Ritualized or habitual practices around eating, greeting, clothing, interaction}
    \item \textit{Moral or symbolic underpinnings of routine social expectations}
    \item \textit{Everyday behavioral norms that guide what is appropriate, respectful, or inappropriate}
    \item \textit{Local variation, but aim for core practices that are widely shared within the group}
\end{itemize}
\end{quote}

\paragraph{\textbf{User Prompt Template}}

For each domain and subcategory, the following template was used:

\begin{quote}
\small
\textit{Please analyze the following OCM entry and extract 8–10 culturally grounded cultural commonsense reasoning topics:}

\textbf{Category:} \textit{[Category Name]}

\textbf{Subcategory:} \textit{[Subcategory Name]}

\textbf{Definition:} \textit{[OCM Definition Text]}

\textit{Focus only on social knowledge that helps people function appropriately in their cultural environment.}
\end{quote}

\paragraph{\textbf{Example Application}}

As an illustration, for the Education category with the Students subcategory:

\begin{quote}
\footnotesize
\textbf{Category:} Education

\textbf{Subcategory:} Education System

\textbf{Definition:} Degree of development and elaboration of formal education; prevalent types of educational specialization (e.g., schools, tutors, apprenticeship); source of support of teachers and educational institutions (e.g., fees from students, ecclesiastical aid, private gifts and endowments); systematization of education (e.g., local schools boards, state educational agencies, voluntary organizations of educational administrators); degree of standardization as to levels, policies, language, and curricula; primary objectives of formal education (e.g., piety, morality, citizenship, vocational skills, intellectual leadership); diffusion of education (e.g., educational statistics); attitudes toward education; etc.
\end{quote}

\clearpage
\subsubsection{Generated and Selected Topics}
\label{app:all_topics}

The following tables present all generated topics organized by domain and subcategory. Selected topics are marked with \sel.

\begin{minipage}{\textwidth}
\centering
\small
\begin{tabular}{p{0.28\textwidth} p{0.65\textwidth}}
\toprule
\textbf{Subcategory} & \textbf{Generated Topics} \\
\midrule

\textbf{Visiting and hospitality (574)} & 
Norms of Reciprocal Visiting $\cdot$ \sel Etiquette in Reception of Visitors $\cdot$ Informal vs Formal Hospitality $\cdot$ Offering of Food and Drink $\cdot$ Social Etiquette at Feasts and Parties $\cdot$ \sel Frequency and Occasions for Visiting $\cdot$ Hospitality Duties and Expectations $\cdot$ Rules of Partaking in Hospitality $\cdot$ Significance of Visiting in Social Relationships \\
\midrule

\textbf{Gift giving (431)} & 
\sel Gift Giving Etiquette $\cdot$ Role of Gifts in Social Relationships $\cdot$ Gift Reciprocity Expectations $\cdot$ \sel Ceremonial Gift Giving $\cdot$ Role of Gifts in Economic Distribution $\cdot$ Recipient's Rights and Privileges $\cdot$ Donor's Rights and Privileges $\cdot$ Gift-Giving on Secular Holidays $\cdot$ Potlatch Practices $\cdot$ Frequency of Gift Giving \\
\midrule

\textbf{Etiquette (576)} & 
\sel Greeting and Salutation Etiquette $\cdot$ Expression of Respect through Obeisances $\cdot$ Complimenting Appropriately $\cdot$ \sel Eating, Drinking, and Smoking Etiquette $\cdot$ Visiting and Travel Etiquette $\cdot$ Deference to Status Superiors $\cdot$ Noblesse Oblige Responsibilities $\cdot$ Social Sanctions for Breaches of Etiquette \\
\bottomrule
\end{tabular}
\captionof{table}{Interpersonal Relations (570)}
\label{app_tab:topics_interpersonal}
\end{minipage}

\begin{minipage}{\textwidth}
\centering
\small
\begin{tabular}{p{0.28\textwidth} p{0.65\textwidth}}
\toprule
\textbf{Subcategory} & \textbf{Generated Topics} \\
\midrule

\textbf{Rest days and holidays (527)} & 
Observance of Rest Days $\cdot$ \sel Conceptualization of Holidays $\cdot$ \sel Secular Festival Practices $\cdot$ \sel Commemoration of Personal Milestones $\cdot$ Holiday Activity Norms $\cdot$ Patriotic Festival Observance $\cdot$ \sel Religious Taboos on Holidays $\cdot$ Norms for Economic Activities on Holidays $\cdot$ Festive Home Visit Etiquette $\cdot$ Holiday Feasting Traditions \\
\midrule

\textbf{Ritual (788)} & 
\sel Symbolic Act Performance $\cdot$ \sel Ritual Gestures $\cdot$ Recital of Religious Formulas $\cdot$ Prescribed Prayer Practices $\cdot$ Mantra and Liturgy Recitation $\cdot$ Participation in Formal Processions $\cdot$ \sel Pilgrimage Practices $\cdot$ Ritual Etiquette and Decorum $\cdot$ Ritual Preparation and Execution $\cdot$ Recognition and Understanding of Ritual Symbols \\
\midrule

\textbf{Organized ceremonial (796)} & 
Recognizing Religious Holidays and Festivals $\cdot$ Observing Rest Days $\cdot$ Participating in Ritual Ceremonies $\cdot$ Understanding Ceremonial Attire and Paraphernalia $\cdot$ Sequencing of Ritual Rites $\cdot$ Interpreting Ritual Symbolism $\cdot$ \sel Engaging with Religious Music and Dance $\cdot$ Attending Religious Services $\cdot$ \sel Timing of Ceremonies $\cdot$ Understanding Religious Dramas and Spectacles \\
\bottomrule
\end{tabular}
\captionof{table}{Festivals and Rituals}
\label{app_tab:topics_festivals}
\end{minipage}

\begin{minipage}{\textwidth}
\centering
\small
\begin{tabular}{p{0.28\textwidth} p{0.65\textwidth}}
\toprule
\textbf{Subcategory} & \textbf{Generated Topics} \\
\midrule

\textbf{Streets and traffic (363)} & 
\sel Understanding Local Traffic Regulations $\cdot$ Familiarity with Street Layout and Design $\cdot$ Comprehending Urban Traffic Composition $\cdot$ Awareness of Street Maintenance Practices $\cdot$ Acknowledging Parking Regulations $\cdot$ Respecting Pedestrian Rights $\cdot$ \sel Adapting to Local Transportation Modes $\cdot$ Navigating Urban Paving and Infrastructure $\cdot$ Recognizing Animal Involvement in Traffic \\
\midrule

\textbf{Transportation (489)} & 
Understanding Public Transportation Systems $\cdot$ Comprehending Transportation Regulations $\cdot$ \sel Carrying Capacity of Transport $\cdot$ Navigating Traffic Volume $\cdot$ Etiquette on Public Transport $\cdot$ \sel Transport during Special Events $\cdot$ Transport System Literacy $\cdot$ Dealing with Transport Delays $\cdot$ Ticketing and Fare Procedures $\cdot$ Environmentally Conscious Transport Choices \\
\bottomrule
\end{tabular}
\captionof{table}{Traffic and Transport Behavior}
\label{app_tab:topics_transport}
\end{minipage}

\begin{table*}[h]
\centering
\small
\begin{tabular}{p{0.28\textwidth} p{0.65\textwidth}}
\toprule
\textbf{Subcategory} & \textbf{Generated Topics} \\
\midrule

\textbf{Education system (871)} & 
Educational Institutions' Financial Support $\cdot$ \sel Formal Educational Structure $\cdot$ Education System Standardization $\cdot$ Objectives of Formal Education $\cdot$ \sel Attitudes Toward Education $\cdot$ Educational Specialization Types $\cdot$ Diffusion of Education $\cdot$ Development of Formal Education \\
\midrule

\textbf{Teachers (875)} & 
Social Status and Respect for Teachers $\cdot$ \sel Norms for Interacting with Teachers $\cdot$ Understanding Teacher Selection Processes $\cdot$ Teacher Salaries and Livelihood $\cdot$ Teacher Tenure and Advancement $\cdot$ Teachers' Academic Freedom $\cdot$ Teacher Associations and Unions $\cdot$ Teachers' Roles Beyond the Classroom $\cdot$ Understanding Teachers' Qualifications and Specializations $\cdot$ Teacher Training and Proficiency Expectations \\
\midrule

\textbf{Students (877)} & 
Composition of Student Body $\cdot$ Student Organizations and Leadership $\cdot$ Social Status of Students $\cdot$ Student Financial Support $\cdot$ Academic Freedom for Students $\cdot$ \sel Student Extracurricular Activities $\cdot$ Student Living and Dining Accommodations $\cdot$ Student-Community Relations $\cdot$ Student-Faculty Relationships $\cdot$ Student Group Values and Behaviors \\
\bottomrule
\end{tabular}
\caption{Education}
\label{app_tab:topics_education}
\end{table*}

\begin{table*}[!h]
\centering
\small
\begin{tabular}{p{0.28\textwidth} p{0.65\textwidth}}
\toprule
\textbf{Subcategory} & \textbf{Generated Topics} \\
\midrule

\textbf{Special garments (292)} & 
\sel Special Occasion Clothing $\cdot$ Weather-Dependent Clothing $\cdot$ Swimwear Etiquette $\cdot$ \sel Headgear and Footwear Norms $\cdot$ Costume Associated Statuses $\cdot$ Activity-Specific Attire $\cdot$ Methods of Wearing Garments $\cdot$ Clothing and Symbolic Meaning $\cdot$ Dress Code for Rituals $\cdot$ Role-Defined Attire \\
\midrule

\textbf{Ornament (301)} & 
\sel Ornamental Attire and Status Indication $\cdot$ Gender Differences in Ornamental Attire $\cdot$ Cultural Materials for Ornament Manufacture $\cdot$ Age-Based Norms for Ornament Wearing $\cdot$ \sel Occasions for Wearing Specific Ornaments $\cdot$ Attachment Modes of Different Ornaments $\cdot$ Types of Ornaments and Their Significance $\cdot$ Etiquette of Gifting Ornaments $\cdot$ Cultural Interpretation of Ornaments \\

\bottomrule
\end{tabular}
\caption{Clothing and Adornment (290 and 300)}
\label{tab:topics_clothing}
\end{table*}

\begin{table*}[!h]
\centering
\small
\begin{tabular}{p{0.28\textwidth} p{0.65\textwidth}}
\toprule
\textbf{Subcategory} & \textbf{Generated Topics} \\
\midrule

\textbf{Food preparation (252)} & 
\sel Food Preparation Techniques $\cdot$ Cultural Cooking Methods $\cdot$ Use of Food Preparation Apparatus $\cdot$ Utilization of Cooking Utensils $\cdot$ \sel Cultural Recipes and Ingredients $\cdot$ Food Preparation Associated Beliefs $\cdot$ Practices Around Commercialized Food Preparation $\cdot$ Social Norms in Food Processing $\cdot$ Rituals in Food Preparation \\
\midrule

\textbf{Diet (262)} & 
\sel Staple Food Consumption $\cdot$ \sel Seasonal Diet Modifications $\cdot$ Dietary Proportions $\cdot$ Group-Specific Dietary Practices $\cdot$ Edible and Harmful Food Discrimination $\cdot$ Food Preferences and Avoidances $\cdot$ Cultural Food Taboos $\cdot$ Gender-Based Food Practices $\cdot$ Social Class and Dietary Norms $\cdot$ Age-Appropriate Foods \\

\bottomrule
\end{tabular}
\caption{Food Processing and Consumption (250 and 260)}
\label{app_tab:topics_food}
\end{table*}

\begin{table*}[h]
\centering
\small
\begin{tabular}{p{0.28\textwidth} p{0.65\textwidth}}
\toprule
\textbf{Subcategory} & \textbf{Generated Topics} \\
\midrule

\textbf{Gestures \& Signs (201)} & 
\sel Social Expression of Emotions $\cdot$ Affirmative and Negative Gestures $\cdot$ Gesture-Based Communication of Size and Shape $\cdot$ Directive Signs in Cultural Communication $\cdot$ Understanding Intertribal Sign Languages $\cdot$ Cultural Nuances in Gesture Interpretation $\cdot$ \sel Non-Verbal Expression of Respect or Disrespect $\cdot$ Social Consequences of Misinterpreted Gestures $\cdot$ Role of Gestures in Conflict Resolution $\cdot$ Gestures Signifying Community Inclusion \\
\midrule

\textbf{Dissemination of News (203)} & 
Understanding Informal Verbal Transmission $\cdot$ Role of Criers and Heralds $\cdot$ Interpreting Bulletins and Newsletters $\cdot$ \sel Navigating the ``Grapevine'' $\cdot$ Etiquette of Sharing News Informally $\cdot$ \sel Trustworthiness of Information Sources $\cdot$ Reaction to Disseminated Information $\cdot$ Responsibility in Information Sharing $\cdot$ Cultural Significance of News Dissemination \\
\bottomrule
\end{tabular}
\caption{Communication (200)}
\label{app_tab:topics_communication}
\end{table*}

\begin{table*}[h]
\centering
\small
\begin{tabular}{p{0.28\textwidth} p{0.65\textwidth}}
\toprule
\textbf{Subcategory} & \textbf{Generated Topics} \\
\midrule

\textbf{Credit (452)} & 
Understanding Credit Extension Practices $\cdot$ \sel Negotiating Credit Advances and Discounts $\cdot$ Credit Ratings and Their Impact $\cdot$ Non-Banking Credit Institutions $\cdot$ \sel Navigating Installment Buying $\cdot$ Role of Bills of Exchange $\cdot$ Managing Credit Terms and Conditions $\cdot$ Culturally Appropriate Credit Communication \\
\midrule

\textbf{Saving \& Investment (454)} & 
Cultural Norms Around Saving $\cdot$ \sel Safekeeping of Valuables $\cdot$ \sel Preferable Investment Forms $\cdot$ Norms Around Investment Banking $\cdot$ Role of Specialized Financial Personnel $\cdot$ Financial Regulation Awareness $\cdot$ Cultural Practices Around Underwriting $\cdot$ Norms of Using Savings Accounts $\cdot$ Principles of Hoarding Wealth \\
\bottomrule
\end{tabular}
\caption{Finance (450)}
\label{app_tab:topics_finance}
\end{table*}

\clearpage

\subsubsection{Seed Question Prompting Details}
\label{app:seed_q_prompts}
This section documents the prompt configuration used for seed question generation.

\paragraph{\textbf{Model Configuration}}

\begin{itemize}
    \item \textbf{Model:} GPT-4-0613
    \item \textbf{Temperature:} 0.7
    \item \textbf{Top-p:} 1.0
\end{itemize}

\paragraph{\textbf{System Prompt}}

The following system prompt was used:

\begin{quote}
\small
\textit{You are a culturally aware commonsense reasoning assistant.}

\textit{Your task is to generate culturally grounded, realistic questions that reflect everyday social norms, expectations, or interactions within a specific region.}

\textit{Each question should:}
\begin{itemize}
    \item \textit{Be relevant to the provided category, subcategory, and topic definition}
    \item \textit{Reflect what someone in that culture is expected to know or understand}
    \item \textit{Avoid trivia, preferences, or niche edge cases}
    \item \textit{Be open ended and lack specific nouns or indicators for the region}
    \item \textit{Must begin with ``In your region''}
    \item \textit{Take inspiration from the example questions provided for that topic}
    \item \textit{Be usable in a cultural commonsense benchmark}
\end{itemize}

\textit{Do not include the answer. Focus on what a culturally competent person should ask or understand in social settings.}
\end{quote}

\paragraph{\textbf{User Prompt Template}}

For each topic, the following template was instantiated with specific category, subcategory, topic, and region information:

\begin{quote}
\small
\textbf{Category:} \textit{[Category Name]}

\textbf{Subcategory:} \textit{[Subcategory Name]}

\textbf{Topic Label:} \textit{[Topic Label]}

\textbf{Topic Definition:} \textit{[Topic Definition]}

\textbf{Region:} \textit{[Target Region]}

\textbf{Question Examples for this Topic:}

\textit{[List of 3-8 example questions demonstrating the style and scope]}

\textit{Generate 20-25 culturally grounded commonsense questions that conform to this topic definition and [Region], like the example questions.}

\textit{Only output the questions, no bullet points, no commentary.}
\end{quote}

\paragraph{\textbf{Example Application}}

As an illustration, for generating questions about visitor reception etiquette in India:

\begin{quote}
\footnotesize
\textbf{Category:} Interpersonal Relations

\textbf{Subcategory:} Visiting and Hospitality

\textbf{Topic Label:} Etiquette in Reception of Visitors

\textbf{Topic Definition:} The traditional norms and behaviors associated with receiving and entertaining visitors in a culturally appropriate manner.

\textbf{Region:} India

\textbf{Question Examples for this Topic:}

\begin{enumerate}
    \item In your region, what is the first thing you do when you enter someone's house? Focus on actions and not greetings.
    \item In your region, what is a traditional drink, aside from water, that is offered to a guest when they visit you? Be as specific as possible.
    \item In your region, what special food items are made when relatives come from out of town?
    \item In your region, how do you traditionally prepare your house for the arrival of guests?
    \item In your region, what are the utensils used to serve meals to guests? Are there any changes made from everyday utensils or are different traditional utensils used?
    \item In your region, what are the common customs or expectations when an out-of-town guest arrives, such as from an airport or train station, in terms of how they travel to your home?
\end{enumerate}
\end{quote}

\clearpage

\subsubsection{Seed Questions for each Topic}
\label{app:seed_q}
\begin{minipage}{\textwidth}
\centering
\small
\begin{tabular}{p{0.28\textwidth} p{0.65\textwidth}}
\toprule
\textbf{Topic} & \textbf{Seed Questions} \\
\midrule

\textbf{Etiquette in Reception of Visitors} & 
1. In your region, what is the first thing you do when you go to someone's house? Focus on actions and not greetings. \newline
2. In your region, what is a traditional drink, aside from water, that is offered to a guest when they visit you? Be as specific as possible. \newline
3. In your region, what special food items are made when relatives come from out of town? \newline
4. In your region, how do you traditionally prepare your house for the arrival of guests? \newline
5. In your region, what are the utensils used to serve meals to guests? Are there any changes made from everyday utensils or are different traditional utensils used? \newline
6. In your region, what are the common customs or expectations when an out-of-town guest arrives, such as from an airport or train station, in terms of how they travel to your home? \\
\midrule

\textbf{Occasions for Visiting} & 
1. In your region what types of occasions usually prompt people to visit neighbors? \newline
2. In your region, is it common to visit friends or relatives unannounced, or is notice expected? \newline
3. In your region, what is the most common personal occasion for visiting relatives? \newline
4. In your region, what is the most common festival for visiting relatives? \\

\bottomrule
\end{tabular}
\captionof{table}{Seed questions for the topics in Visiting and Hospitality subcategory}
\label{app_tab:topics_visiting_hospitality}
\end{minipage}

\begin{minipage}{\textwidth}
\centering
\small
\begin{tabular}{p{0.28\textwidth} p{0.65\textwidth}}
\toprule
\textbf{Topic} & \textbf{Seed Questions} \\
\midrule

\textbf{Gift Giving Etiquette} & 
1. In your region, what are the most common occasions for giving gifts (e.g., festivals, weddings, birthdays, housewarmings, visits)? \newline
2. In your region, when receiving a gift, is it generally expected to open it immediately in front of the giver, or to open it later? \newline
3. In your region, what types of gifts are considered universally acceptable for common occasions? \newline
4. In your region, how do cultural or religious beliefs influence the choice of gifts or the manner of giving them (e.g., avoiding certain numbers, colors)? \newline
5. In your region, what are considered inappropriate or unlucky gifts to give someone? \newline
6. In your region, what is the social expectation when someone receives a gift, what do they say or do? \\
\midrule

\textbf{Ceremonial Gift Giving} & 
1. In your region, what are the commonly accepted types of gifts for a wedding ceremony? \newline
2. In your region, during a baby naming ceremony or a child’s first birthday, what kind of gifts are typically given to the child or the parent? \newline
3. In your region, when attending a religious function at someone's home, is it customary to bring a gift, and if so, what types are appropriate and most common? \newline
4. In your region, is there a custom of giving gifts to priests or religious officiants during ceremonies, and what form do these gifts usually take? \newline
5. In your region, are gifts during religious or spiritual events expected to be new, handmade, or of a particular material? \newline
6. In your region, are there gifts that must not be given during certain ceremonies due to religious or cultural taboos? If so, mention the item and the occasion. \\

\bottomrule
\end{tabular}
\captionof{table}{Seed questions for the topics in Gift Giving subcategory}
\label{app_tab:topics_gift_giving}
\end{minipage}

\begin{table*}[h]
\centering
\small
\begin{tabular}{p{0.28\textwidth} p{0.65\textwidth}}
\toprule
\textbf{Topic} & \textbf{Seed Questions} \\
\midrule

\textbf{Greeting and Salutation Etiquette} & 
1. In your region, what are the most common verbal greetings used when meeting someone for the first time, both formally and informally? \newline
2. In your region, what type of physical gestures (like bowing, touching feet, or handshakes) accompany greetings? \newline
3. In your region, what type of salutation is used when addressing someone of high status or authority? \newline
4. In your region, what type of greeting is expected when entering a religious or spiritual place? \newline
5. In your region, what are the customary ways to bid farewell to someone, both in formal and informal situations? \\
\midrule

\textbf{Eating, Drinking, and Smoking Etiquette} & 
1. In your region, what are the customary practices for beginning and ending a meal, especially in a family or formal setting? \newline
2. In your region, where is it generally considered acceptable to smoke (e.g., designated areas, private homes), and where is it strictly prohibited or frowned upon? \newline
3. In your region, is there any type of seating arrangement that is common during formal or family meals? \newline
4. In your region, what type of food is considered inappropriate to refuse? \newline
5. In your region, what type of hand (left or right) is traditionally used for eating, and why? \\

\bottomrule
\end{tabular}
\caption{Seed questions for the topics in Etiquette subcategory}
\label{app_tab:topics_etiquette}
\end{table*}

\begin{table*}[h]
\centering
\small
\begin{tabular}{p{0.28\textwidth} p{0.65\textwidth}}
\toprule
\textbf{Topic} & \textbf{Seed Questions} \\
\midrule

\textbf{Conceptualization of Holidays} & 
1. In your region, what is the most anticipated and widely celebrated festival and its significance or purpose? \newline
2. In your region, what is the most anticipated and widely celebrated patriotic holiday and its significance or purpose? \newline
3. In your region, what are some other holidays/festivals that are celebrated and the cultural significance of them? \\
\midrule

\textbf{Secular Festival Practices} & 
1. In your region, what are the typical decor activities and practices associated with a housewarming celebration? \newline
2. In your region, what are the typical decor activities and practices associated with harvest celebrations? \newline
3. In your region, what are the decor typical activities and practices associated with the most widely celebrated festival? \newline
4. In your region, what month is your harvest festival celebrated? \\
\midrule

\textbf{Commemoration of Personal Milestones} & 
1. In your region, what is the first big moment celebrated after your child's birth? \newline
2. In your region, what are the most important birthdays in someone's life? \newline
3. In your region, are there any cultural norms associated with the first anniversary of a couple? If any, what are they? Focus on any celebrations or cultural customs that might be performed. \newline
4. In your region, what other personal milestones of an individual are commonly celebrated apart from anniversaries and birthdays? \\
\midrule

\textbf{Religious Taboos on Holidays} & 
1. In your region, are there any festivals or holidays that have restrictions on activities performed, for example food restrictions or action restrictions? If any, name them and also the restrictions. \newline
2. In your region, are there any days of the week that hold certain constraints, for example food or activity related restrictions? If any, name the day and also the restriction. \newline
3. In your region, are there any hygiene restrictions that hold during certain festivals or holidays? If yes, name the holiday and the restrictions. \\

\bottomrule
\end{tabular}
\caption{Seed questions for the topics in Rest Days and Holidays subcategory}
\label{app_tab:topics_holidays}
\end{table*}

\begin{table*}[h]
\centering
\small
\begin{tabular}{p{0.28\textwidth} p{0.65\textwidth}}
\toprule
\textbf{Topic} & \textbf{Seed Questions} \\
\midrule

\textbf{Symbolic Act Performance} & 
1. In your region, what types of lamps are lit (if any) and what is their significance? \newline
2. In your region, what is a specific ritual that you perform often and what are the specific actions you perform to observe that ritual? \newline
3. In your region, what is often offered to deities? Be as specific as possible. \newline
4. In your region, what are the specific actions you perform when visiting a religious institution? \newline
5. In your region, when receiving spiritual blessings, what physical and verbal responses are customary? \\
\midrule

\textbf{Ritual Gestures} & 
1. In your region, what is the customary gesture for showing reverence to a sacred text, and why is it performed? \newline
2. In your region, what is the most common gesture done during a religious ritual? \newline
3. In your region, what is the customary gesture for offering food to a deity, and how is it performed? \newline
4. In your region, what is a gesture associated with remembrance or honor of a religion? \newline
5. In your region, is there a gesture to show respect after touching someone or something with your feet? \\
\midrule

\textbf{Pilgrimage Practices} & 
1. In your region, what preparatory practices like fasting or special clothing precede important pilgrimages? \newline
2. In your region, what rituals are performed immediately upon reaching a pilgrimage destination? \newline
3. In your region, what sacred items do pilgrims carry back from their destination, and how are these used later? \newline
4. In your region, what is the most popular pilgrimage? \newline
5. In your region, who can undertake pilgrimages? \newline
6. In your region, what changes in movement or attire occur during certain parts of a pilgrimage, and what might these signify? \\

\bottomrule
\end{tabular}
\caption{Seed questions for the topics in Ritual subcategory}
\label{app_tab:topics_ritual}
\end{table*}

\begin{table*}[h]
\centering
\small
\begin{tabular}{p{0.28\textwidth} p{0.65\textwidth}}
\toprule
\textbf{Topic} & \textbf{Seed Questions} \\
\midrule

\textbf{Engaging with Religious Music and Dance} & 
1. In your region, what is the traditional dance form associated with your culture or rituals? \newline
2. In your region, what is the traditional music form associated with your culture or rituals? \newline
3. In your region, are there any specific dances performed during specific rituals, if yes, specify the dance as well as the ritual. \newline
4. In your region, is there specific music that is played during specific rituals, if yes, specify the music as well as the ritual. \newline
5. In your region, who typically participates in these traditional dance forms or who is it performed by? \\
\midrule

\textbf{Timing of Ceremonies} & 
1. In your region, how is an auspicious time for important ceremonies, such as weddings, typically chosen, and who is involved in that decision? \newline
2. In your region, how do spiritual advisors determine the right time to begin a new venture or embark on a journey? \newline
3. In your region, how do agricultural rhythms and seasonal changes influence the timing of festivals, rituals, or community ceremonies related to the land? \newline
4. In your region, are certain days or times avoided for house-related rituals, and what beliefs influence those choices? \newline
5. In your region, are there certain periods during the year when major activities are paused or avoided, and what is the reasoning behind this? \newline
6. In your region, is there a specific calendar followed that is not typical to the calendar in the rest of the world, if yes, specify it. \newline
7. In your region, are there certain timings avoided to take flights or journeys? If yes, please specify. \newline
8. In your region, are there certain times of the year where there are food restrictions, if yes, specify.  \\

\bottomrule
\end{tabular}
\caption{Seed questions for the topics in Organized Ceremonial subcategory}
\label{app_tab:topics_ceremonial}
\end{table*}

\begin{table*}[h]
\centering
\small
\begin{tabular}{p{0.28\textwidth} p{0.65\textwidth}}
\toprule
\textbf{Topic} & \textbf{Seed Questions} \\
\midrule

\textbf{Understanding Local Traffic Regulations} & 
1. In your region, is there a designated spot where pedestrians normally cross the road? If not, specify where and how the pedestrians normally cross the road. \newline
2. In your region, if not in a residential area, where are the cars normally parked? Is there any payment associated with this typical parking norm? \newline
3. In your region, what are the conventions and frequencies of honking? Focus on how often people honk and for what reasons. \newline
4. In your region, what side of the road to people normally drive on?  \\
\midrule

\textbf{Adapting to Local Transportation Modes} & 
1. In your region, what is the most widely used form of public/local transportation? \newline
2. In your region, what is the most unique form of local transportation that may not be found in other regions? \newline
3. In your region, are the public/local transportation methods used to transport any other goods or services apart from people? If yes, please specify the mode of transport as well as the goods/services transported. \newline
4. In your region, what are the typical occasions to use public transport? Focus on where people generally commute to using local transportation. \newline
5. In your region, do people use digital tools or local networks to stay informed about delays, route changes, or real-time public transport updates? If so specify the tool as well as the public transport it is used for. \newline
6. In your region, what are the typical apps used to book public transport routes? Specify the app as well as the public transport. \newline
7. In your region, what payment options are commonly used for buses, trains, or other public transport, and which ones are most convenient for daily users? \\

\bottomrule
\end{tabular}
\caption{Seed questions for the topics in Streets and Traffic subcategory}
\label{app_tab:topics_traffic_2}
\end{table*}

\begin{table*}[h]
\centering
\small
\begin{tabular}{p{0.28\textwidth} p{0.65\textwidth}}
\toprule
\textbf{Topic} & \textbf{Seed Questions} \\
\midrule

\textbf{Carrying Capacity of Transport} & 
1. In your region, for the most used public transport, is there a carrying capacity maintained normally or is the capacity often broken? \newline
2. In your region, what is the most crowded public transport mode? \newline
3. In your region, what accommodations are made to guarantee more space when reaching carrying capacity? For example, sharing seats with the driver in the auto, etc. Be specific. \\
\midrule

\textbf{Transport during Special Events} & 
1. In your region, are there any new routes or transport added to accommodate for special events like festivals, etc? If any, specify the transportation added and the special event it was added for. \newline
2. In your region, what is the influence on public transport during a major festival, is the use increased or decreased? Specify the mode of transportation as well as the change. \newline
3. In your region, is there commonly a change from your regular transportation method to another during a major festival? If so specify the regular transport, the changed transport, and the major festival. \\

\bottomrule
\end{tabular}
\caption{Seed questions for the topics in Transportation subcategory}
\label{app_tab:topics_transportation}
\end{table*}

\begin{table*}[h]
\centering
\small
\begin{tabular}{p{0.28\textwidth} p{0.65\textwidth}}
\toprule
\textbf{Topic} & \textbf{Seed Questions} \\
\midrule

\textbf{Formal Educational Structure} & 
1. In your region, are there any local school boards followed? If yes, please specify them. \newline
2. In your region, what are the common school boards that children go to? Please specify all the common board names. \newline
3. In your region, what are the most important exams that children give in their education up until high school? \newline
4. In your region, what grades are normally present in a school? Be as specific as possible. \newline
5. In your region, do students normally change schools since some grades are absent in the school? If yes, specify when schools are changed. \newline
6. In your region, what are the typical exams that students give post high school, for university/college purposes? \\
\midrule

\textbf{Attitudes Toward Education} & 
1. In your region, how many levels of education are considered necessary for a child to complete? Focus on levels e.g. high school, college, etc. \newline
2. In your region, are there any streams of education considered more beneficial for the child's future over the others? If yes, please specify the streams. \newline
3. In your region, how important is education to get a job? \newline
4. In your region, what educational board is considered the best to provide education to children? \\

\bottomrule
\end{tabular}
\caption{Seed questions for the topics in Education System subcategory}
\label{app_tab:topics_education_system}
\end{table*}

\begin{table*}[h]
\centering
\small
\begin{tabular}{p{0.28\textwidth} p{0.65\textwidth}}
\toprule
\textbf{Topic} & \textbf{Seed Questions} \\
\midrule

\textbf{Norms for Interacting with Teachers} & 
1. In your region, how are teachers addressed when they enter the classroom? Focus on actions and sayings and be descriptive. \newline
2. In your region, what is the typical status of teachers? Focus on status in terms of the amount of respect. \newline
3. In your region, what is considered the appropriate way to address a teacher? Focus on sayings used. \newline
4. In your region, what actions or gestures are considered disrespectful when interacting with teachers? \\
\midrule

\textbf{Student Extracurricular Activities} & 
1. In your region, what are the typical activities that students engage in outside of school? Think about their after school activities. \newline
2. In your region, is it more common for students to engage in extracurricular activities or further academics after school? \newline
3. In your region, when students go for academic help after school, what does that usually look like? (home tutoring, coaching center. etc) \newline
4. In your region, what are the common extracurricular activities carried out within the school? \\

\bottomrule
\end{tabular}
\caption{Seed questions for the topics in Teachers and Students subcategory}
\label{app_tab:topics_teachers_students}
\end{table*}

\begin{table*}[h]
\centering
\small
\begin{tabular}{p{0.28\textwidth} p{0.65\textwidth}}
\toprule
\textbf{Topic} & \textbf{Seed Questions} \\
\midrule

\textbf{Special Occasion Clothing} & 
1. In your region, what is the traditional outfit of a bride in a wedding ceremony? \newline
2. In your region, what is the traditional color of the bride's outfit? \newline
3. In your region, what is the traditional outfit of a groom in a wedding ceremony? \newline
4. In your region, what is the traditional outfit associated with the harvest festival of your region? Name and describe the outfit as well as the harvest festival. \newline
5. In your region, what is the traditional outfit associated with the most widely celebrated festival of your region? Name and describe the outfit as well as the festival. \newline
6. In your region, what fabrics are seen as auspicious for celebration wear? \newline
7. In your region, what is the traditional attire worn by women if attending a religious ceremony at someone's? Please specify the religious ceremony as well as the attire. \newline
8. In your region, what is the traditional attire worn by men if attending a religious ceremony at someone's? Please specify the religious ceremony as well as the attire. \\

\midrule

\textbf{Headgear and Footwear Norms} & 
1. In your region, what are the occasions of wearing certain head coverings that are not normally worn? For e.g. during religious ceremonies, etc. Be as specific as possible and describe the occasion. \newline
2. In your region, what are the types of head covering, if any, worn during wedding ceremonies by the bride that are not ordinarily worn? \newline
3. In your region, do married woman wear certain head coverings as a signifier of their marriage? If yes, please specify the type of head covering. \newline
4. In your region, what types of head covering, if any, are worn during wedding ceremonies by the groom that are not ordinarily worn? \newline
5. In your region, are there any special types of footwear worn by grooms during wedding ceremonies that are not ordinarily worn? If yes, please specify the type of the footwear. \newline
6. In your region, what are some formal footwear choices for men that are not ordinarily worn? \newline
7. In your region, what are some formal footwear choices for women that are not ordinarily worn? \newline
8. In your region, are there any materials preferred for formal footwear? If yes, please specify the materials. \\
\bottomrule
\end{tabular}
\caption{Seed questions for the topics in Special Garments subcategory}
\label{app_tab:topics_special_garments}
\end{table*}

\begin{table*}[h]
\centering
\small
\begin{tabular}{p{0.28\textwidth} p{0.65\textwidth}}
\toprule
\textbf{Topic} & \textbf{Seed Questions} \\
\midrule

\textbf{Ornamental Attire and Status Indication} & 
1. In your region, are there specific ornaments that are seen as signs of wealth and how do people view them? \newline
2. In your region, what are the ornaments worn during a wedding that signify higher social standing? \newline
3. In your region what are ornaments worn by men that showcase a higher status symbol? \newline
4. In your region what are ornaments worn by women that showcase a higher status symbol? \\
\midrule

\textbf{Occasions for Wearing Specific Ornaments} & 
1. In your region, what type of traditional ornaments are usually worn by the bride during weddings? \newline
2. In your region, what type of traditional ornaments are usually worn by the groom during weddings? \newline
3. In your region, what types of ornaments are typically avoided during mourning or funerals? \newline
4. In your region, are there any ornaments that are specially worn only during certain festivals? If yes, specify the ornament as well as the festival. \newline
5. In your region, are there any ornaments that are avoided when traveling or visiting crowded areas? If yes, specify the ornaments. \\

\bottomrule
\end{tabular}
\caption{Seed questions for the topics in Ornament subcategory}
\label{app_tab:topics_ornament}
\end{table*}

\begin{table*}[h]
\centering
\small
\begin{tabular}{p{0.28\textwidth} p{0.65\textwidth}}
\toprule
\textbf{Topic} & \textbf{Seed Questions} \\
\midrule

\textbf{Food Preparation Techniques} & 
1. In your region what are the traditional tools utilized for grinding spices for food preparation? \newline
2. In your region what are typical methods used for preserving food for long periods? \newline
3. In your region what types of stove or cooking setup is used for making traditional dishes? \newline
4. In your region, how are coconut or other hard ingredients broken and processed for cooking? \newline
5. In your region, what types of vessels are more commonly used for everyday cooking? \newline
6. In your region are there traditional condiments made at home? If yes, what are they? \\
\midrule

\textbf{Cultural Recipes and Ingredients} & 
1. In your region, what is the featured spice in most of your traditional food items? \newline
2. In your region, what is the meat (if any) associated with most of your traditional dishes? \newline
3. In your region, what is a traditional vegetable that is featured in most traditional dishes? \newline
4. In your region, what is the traditional dish associated with your most celebrated festival? Name the festival as well as the dish. \newline
5. In your region, is there a food ceremony normally associated with any of your festivals? If yes, please specify the name of the food ceremony as well as the festival. \newline
6. In your region, are any specific condiments used during special occasions? \newline
7. In your region, are any specific condiments avoided during certain occasions? \newline
8. In your region are there condiments people avoid mixing together? If yes, specify. \\

\bottomrule
\end{tabular}
\caption{Seed questions for the topics in Food Preparation subcategory}
\label{app_tab:topics_food_preparation}
\end{table*}

\begin{table*}[h]
\centering
\small
\begin{tabular}{p{0.28\textwidth} p{0.65\textwidth}}
\toprule
\textbf{Topic} & \textbf{Seed Questions} \\
\midrule

\textbf{Staple Food Consumption} & 
1. In your region, what is considered a staple food source? \newline
2. In your region, what type of grain or cereal is usually served with lunch or dinner? \newline
3. In your region, what type of staple food is usually eaten during fasting periods? \newline
4. In your region, what are the typical breakfast staples? \newline
5. In your region, what is the most common accompaniment served with rice? \newline
6. In your region, what is the most commonly used bread in most dishes? \newline
7. In your region, after a main meal, what is commonly consumed as a digestive aid or mouth freshener? \\
\midrule

\textbf{Seasonal Diet Modifications} & 
1. In your region, what special dishes are prepared during the rainy season? \newline
2. In your region, what changes are made to the diet during winters? \newline
3. In your region, what traditional drinks are preferred during summers? \newline
4. In your region, what traditional drinks are preferred during winters? \newline
5. In your region, are any foods that are avoided in certain seasons? If yes, specify the food as well as the season. \\

\bottomrule
\end{tabular}
\caption{Seed questions for the topics in Diet subcategory}
\label{app_tab:topics_diet}
\end{table*}

\begin{table*}[h]
\centering
\small
\begin{tabular}{p{0.28\textwidth} p{0.65\textwidth}}
\toprule
\textbf{Topic} & \textbf{Seed Questions} \\
\midrule

\textbf{Social Expression of Emotions} & 
1. In your region how do people express disagreement or disapproval without using any words? \newline
2. In your region what hand gestures are considered rude or insulting? \newline
3. In your region what gestures are used to greet someone warmly versus formally? \newline
4. In your region how do people express gratitude nonverbally? \newline
5. In your region what gestures or signs express anger or frustration? \\
\midrule

\textbf{Non-Verbal Expression of Respect or Disrespect} & 
1. In your region, how do you greet an elder? Focus on gestures. \newline
2. In your region, what gestures signify respect towards teachers? \newline
3. In your region, what gestures are performed to respectfully greet a religious leader? \newline
4. In your region, how do you display respect for the national flag or anthem through gestures? \newline
5. In your region, how do you display respect to a deceased person during a funeral ceremony? Focus on gestures. \\

\bottomrule
\end{tabular}
\caption{Seed questions for the topics in Gestures \& Signs subcategory}
\label{app_tab:topics_gestures}
\end{table*}

\begin{table*}[h]
\centering
\small
\begin{tabular}{p{0.28\textwidth} p{0.65\textwidth}}
\toprule
\textbf{Topic} & \textbf{Seed Questions} \\
\midrule

\textbf{Navigating the ``Grapevine''} & 
1. In your region, what is the typical informal digital source used to convey news and information? \newline
2. In your region, what role do tea shops, local markets, or hair salons play in spreading informal news? \newline
3. In your region, what are the typical practices done to avoid spreading sensitive information on the ``grapevine''? \newline
4. In your region, are digital platforms used to resolve or discuss issues related to the neighborhood or community? If yes specify the platform as well as the issues discussed through it. \newline
5. In your region, do religious institutions use any digital platforms to spread information or keep people aware of religious practices? If yes, specify the digital platform, religion, and the information spread through it. \\
\midrule

\textbf{Trustworthiness of Information Sources} & 
1. In your region, what are the formal digital news platforms that are considered reliable? \newline
2. In your region, what are the informal digital news platforms that are considered reliable? \newline
3. In your region, do people still commonly continue to get their news through newspapers? \newline
4. In your region, what are the most common newspapers that are considered reliable? \newline
5. In your region, are there any magazines that kids normally read? If yes, specify. \\

\bottomrule
\end{tabular}
\caption{Seed questions for the topics in Dissemination of News and Information subcategory}
\label{app_tab:topics_news}
\end{table*}

\begin{table*}[h]
\centering
\small
\begin{tabular}{p{0.28\textwidth} p{0.65\textwidth}}
\toprule
\textbf{Topic} & \textbf{Seed Questions} \\
\midrule

\textbf{Negotiating Credit Advances and Discounts} & 
1. In your region, is it common to negotiate prices before buying, or is the price usually fixed? If so specify for what purchases. \newline
2. In your region, are people more likely to offer discounts to familiar customers or strangers? If so specify for what purchases. \newline
3. In your region, what kinds of purchases are most commonly negotiated for better rates? \newline
4. In your region, do certain professions or groups receive more flexibility with credit or discounts? If so specify. \newline
5. In your region, do people expect discounts when buying in bulk? If so specify. \newline
6. In your region, are there particular times of the year or specific events when discounts and promotional offers are more prevalent? If so, specify. \newline
7. In your region, what kinds of services or professional fees are typically open to negotiation for better rates? \\
\midrule

\textbf{Navigating Installment Buying} & 
1. In your region, what kinds of products are most commonly purchased through installment plans? \newline
2. In your region, are people more likely to use formal (bank/store) or informal (person-to-person) installment agreements? \newline
3. In your region, how do family or community opinions influence a person's decision to opt for installment buying? \newline
4. In your region what are common schemes offered for installment plans (eg: no cost emi, guarantor etc) \newline
5. In your region, how do shopkeepers decide whether a customer is eligible for paying in installments? \newline
6. In your region, are certain professions more likely to use installment buying than others? \\

\bottomrule
\end{tabular}
\caption{Seed questions for the topics in Credit subcategory}
\label{app_tab:topics_credit}
\end{table*}

\begin{table*}[h]
\centering
\small
\begin{tabular}{p{0.28\textwidth} p{0.65\textwidth}}
\toprule
\textbf{Topic} & \textbf{Seed Questions} \\
\midrule

\textbf{Safekeeping of Valuables} & 
1. In your region, do people commonly keep their important valuables at home? If not, specify where they keep it. \newline
2. In your region, how do people protect valuable things when they travel? \newline
3. In your region, do people prefer keeping cash at home or in a bank, and why? \newline
4. In your region, is there a strong cultural emphasis on keeping precious metals within the household for generations, and how is it typically protected? \newline
5. In your region, how do cultural beliefs or superstitions sometimes influence the placement or display of certain valuables within a home? \newline
6. In your region, what items do people typically hoard as a way of preserving wealth? \\
\midrule

\textbf{Preferable Investment Forms} & 
1. In your region, what is considered a preferred long-term investment for families? (Like gold, land etc) \newline
2. In your region, how do religious or cultural beliefs influence what people invest in? \newline
3. In your region, what kind of property is most commonly purchased as an investment? \newline
4. In your region, what kind of items are bought as ``prestige'' investments beyond financial value? \newline
5. In your region, do people invest in religious donations or temples as a spiritual investment? If yes, specify. \newline
6. In your region, how do marriage customs influence what type of assets families prioritize? \\

\bottomrule
\end{tabular}
\caption{Seed questions for the topics in Saving \& Investment subcategory}
\label{app_tab:topics_saving_investment}
\end{table*}

\clearpage
\subsection{Study Details}
\label{app:study_details}

\subsubsection{Participant Criteria}
Participants were required to have lived in their target region for more than half their lifetime.

\subsubsection{Study Design}
Each participant completed 41 forms, with each form containing a maximum of 15 questions (611 total questions divided across forms). To ensure participants provided region-specific responses, each question was prefaced with ``In your region.'' For example, the question ``Are gifts opened in front of the giver or later?'' was presented as ``In your region, are gifts opened in front of the giver or later?'' This framing served as a consistent reminder for participants to draw upon their local cultural knowledge. The form interface is available at: \url{https://cultural-survey-frontend.vercel.app/}.

Participants responded to questions in their own words, allowing us to capture natural cultural knowledge rather than forcing responses into predetermined categories. Each form took between 20--67 minutes to complete. We implemented various attention checks throughout, and responses were reviewed and scanned for any AI-generated content.

\subsubsection{Compensation}
Participants were compensated at \$8.00 per hour, aligning with Prolific's fair payment standards.

\subsubsection{Ethics and Consent}
This study received approval from our institution's Research Ethics Board (REB). All participants provided informed consent before beginning the study, acknowledging data usage, anonymization procedures, and their right to withdraw at any time.

\subsubsection{Data Collection Period}
Responses were collected between October and November 2025.

\clearpage
\subsection{Agreement Validation and Override Analysis}
\label{app:agreement_validation}

GPT-4o provided preliminary classifications for three types of agreement across all 
question-region combinations: intra-regional consensus (whether 4--5 participants 
within a region agreed), inter-regional agreement (whether two regions shared the 
same practice), and universal agreement (whether all five regions agreed).

Two authors independently reviewed all cases using a custom annotation tool displaying: 
(1) the question, (2) GPT-4o's preliminary assessment, and (3) all participant responses. 
For each case, annotators decided whether responses were semantically equivalent and, 
if so, established the gold standard answer. Inter-annotator agreement between the two 
human annotators was perfect (Fleiss' $\kappa = 1.0$) across all judgment types.

\subsubsection{Override Rates}

Table~\ref{tab:llm_override_rates} presents the rates at which human annotators 
overrode GPT-4o's preliminary classifications.

\begin{minipage}{\textwidth}
\centering
\small
\begin{tabular}{lcccc}
\toprule
\textbf{Agreement Type} & \textbf{Pairs} & \textbf{Raw LLM} & \textbf{Production} & \textbf{Overrides} \\
\midrule
\textbf{Intra-Regional} & 2,540 & 1,607 (63.3\%) & 1,605 (63.2\%) & 194 (7.6\%) \\
\quad Added (LLM No → Human Yes) & & & & 96 \\
\quad Removed (LLM Yes → Human No) & & & & 98 \\
\midrule
\textbf{Inter-Regional} & 2,197 & 1,931 (87.9\%) & 1,299 (59.1\%) & 636 (28.9\%) \\
\quad Added (LLM No → Human Yes) & & & & 2 \\
\quad Removed (LLM Yes → Human No) & & & & 634 \\
\midrule
\textbf{Universal} & 106 & 75 (70.8\%) & 49 (46.2\%) & 26 (24.5\%) \\
\quad Added (LLM No → Human Yes) & & & & 0 \\
\quad Removed (LLM Yes → Human No) & & & & 26 \\
\bottomrule
\end{tabular}
\captionof{table}{Human override rates of GPT-4o's preliminary agreement classifications. 
Intra-regional overrides were balanced between additions and removals (7.6\% override rate), 
while inter-regional (28.9\%) and universal (24.5\%) cases showed predominantly removals, 
indicating GPT-4o over-identified cross-regional consensus.}
\label{tab:llm_override_rates}
\end{minipage}

\clearpage
\subsection{Intra-Region Agreement Prompting Details}
\label{app:intra_region_agreement}

\subsubsection{Model Configuration}

\begin{itemize}
    \item \textbf{Model:} GPT-4o-2024-08-06
    \item \textbf{Temperature:} 0.1
    \item \textbf{Max Tokens:} 800--3000 (progressive increase across retry attempts)
    \item \textbf{Retry Logic:} Up to 5 retry attempts with increasing token limits (800 $\rightarrow$ 1200 $\rightarrow$ 1500 $\rightarrow$ 2000 $\rightarrow$ 2500 $\rightarrow$ 3000)
\end{itemize}

\subsubsection{System Prompt}

The following system prompt was used:

\begin{quote}
\small
\textit{You are an expert at analyzing cultural agreement. You must respond with valid, complete JSON only, no additional text.}
\end{quote}

\subsubsection{User Prompt Template}

For each question with regional responses, the following user prompt template was used:

\begin{quote}
\small
\textit{Analyze agreement among [N] responses from [REGION] about regional practices.}

\textbf{QUESTION:} \textit{[Question Text]}

\textbf{NUMBERED RESPONSES:}

\textit{Response 1: [Answer 1]}

\textit{Response 2: [Answer 2]}

\textit{...}

\textbf{CORE INSTRUCTION:} \textit{Only look for concepts that directly answer the question asked. This is imperative above all else.}

\textbf{ANALYSIS RULES:}
\begin{enumerate}
    \item \textit{Look for the SAME underlying concept across responses (semantic similarity counts)}
    \item \textit{Spelling variations, spacing differences, and synonyms count as the SAME concept}
    \item \textit{You must quote exact text but recognize when different words mean the same thing}
    \item \textit{[THRESHOLD]+ different responses must mention the same underlying concept}
\end{enumerate}

\textit{where [THRESHOLD] = 4 if N $\geq$ 5, otherwise max(2, N-1)}

\textbf{CONCEPT IDENTIFICATION EXAMPLES:}

\textit{Question: ``What foods are eaten during festivals?''}
\begin{itemize}
    \item \textit{Response mentions ``sweets'' $\rightarrow$ Answers the question}
    \item \textit{Response mentions ``celebration'' $\rightarrow$ Doesn't answer what food}
\end{itemize}

\textit{Question: ``What nonverbal actions are disrespectful to elders?''}
\begin{itemize}
    \item \textit{Response mentions ``pointing'' $\rightarrow$ Answers the question (specific action)}
    \item \textit{Response mentions ``being rude'' $\rightarrow$ Doesn't answer what action}
\end{itemize}

\textit{Question: ``In your region, what is the customary gesture for offering food to a deity, and how is it performed?''}
\begin{itemize}
    \item \textit{Response: ``We offer food to deity's by putting it on a clean plate'' $\rightarrow$ Concept extracted should be ``putting it on a clean plate'' since that is what answers the question}
    \item \textit{Response: ``We offer food to deity's by putting it on a clean plate'' $\rightarrow$ Concept extracted should NOT be ``food is offered'', that is not the answer to the question}
\end{itemize}

\textbf{SEMANTIC MATCHING EXAMPLES:}
\begin{itemize}
    \item \textit{``Raksha Bandhan'' = ``Rakshabandhan'' = ``rakhi'' (same festival)}
    \item \textit{``clean house'' = ``cleaning home'' = ``tidy up house'' (same activity)}
    \item \textit{``Diwali'' = ``Deepawali'' (same festival)}
    \item \textit{``new clothes'' = ``fresh clothing'' = ``new garments'' (same concept)}
\end{itemize}

\textbf{STEP-BY-STEP ANALYSIS:}
\begin{enumerate}
    \item \textit{Analyse if the response is even answering the question before moving on to getting the concepts}
    \item \textit{Extract concepts from each response with exact quotes that answer the question}
    \item \textit{Group semantically similar concepts together (consider spelling, synonyms, variants)}
    \item \textit{Count how many different responses mention each concept group}
    \item \textit{Agreement exists if any concept group appears in [THRESHOLD]+ responses}
\end{enumerate}

\textbf{VERIFICATION FORMAT:}

\textit{For each concept group, show the evidence and reasoning.}
\end{quote}

\subsubsection{Required JSON Output Format}

The model was required to return structured JSON with the following schema:

\begin{quote}
\footnotesize
\begin{verbatim}
{
  "step_by_step_extraction": {
    "response_1_concepts": ["concept from response 1"],
    "response_2_concepts": ["concept from response 2"],
    ...
  },
  "semantic_grouping": {
    "concept_group_name": {
      "responses_and_quotes": {
        "1": "exact quote from response 1",
        "2": "exact quote from response 2",
        ...
      },
      "semantic_explanation": "Why these quotes represent 
                               the same concept",
      "count": N
    }
  },
  "agreement_found": true/false,
  "threshold_met": "X out of Y responses mention the same 
                    underlying concept",
  "common_concepts": [
    {
      "concept": "unified concept name",
      "responses_mentioning": [1, 2, 3],
      "exact_quotes_proof": ["quote 1", "quote 2", "quote 3"],
      "semantic_note": "Explanation of any spelling/synonym 
                        variations"
    }
  ],
  "summary": "Brief explanation recognizing semantic 
              similarity while showing evidence"
}
\end{verbatim}
\end{quote}

\subsubsection{Retry Mechanism}

To handle truncated or incomplete responses, an automatic retry system was implemented:

\begin{itemize}
    \item \textbf{Trigger Conditions:} Empty responses, unparseable JSON, missing required fields, or agreement found with empty concept lists
    \item \textbf{Progressive Token Increase:} Each retry attempt increased \texttt{max\_tokens} (800 $\rightarrow$ 1200 $\rightarrow$ 1500 $\rightarrow$ 2000 $\rightarrow$ 2500 $\rightarrow$ 3000)
    \item \textbf{Maximum Attempts:} 5 retries per question
\end{itemize}

\subsubsection{Agreement Threshold Logic}

The agreement threshold was dynamically calculated based on the number of responses:

\begin{equation}
\text{Threshold} = 
\begin{cases} 
4 & \text{if } N \geq 5 \\
\max(2, N-1) & \text{otherwise}
\end{cases}
\end{equation}

where $N$ is the total number of responses for a given question in a region. In our dataset, all questions had exactly $N = 5$ responses per region, resulting in a consistent agreement threshold of 4 responses. This means that for agreement to be found, at least 4 out of 5 responses (80\%) needed to mention the same underlying concept.

\clearpage

\subsection{Inter-Region Agreement Prompting Details}
\label{app:inter_region_agreement}
\subsubsection{Model Configuration}

\begin{itemize}
    \item \textbf{Model:} GPT-4o-2024-08-06
    \item \textbf{Temperature:} 0.1
    \item \textbf{Max Tokens:} 800
\end{itemize}

\subsubsection{Question Matching Methodology}

Questions were matched between regions using normalized question text rather than question numbers. Text normalization involved:

\begin{itemize}
    \item Converting to lowercase
    \item Removing extra whitespace
    \item Stripping leading/trailing spaces
\end{itemize}

\subsubsection{System Prompt}

The following system prompt was used:

\begin{quote}
\small
\textit{You are an expert at comparing cultural concepts for semantic similarity. You must respond with valid JSON only.}
\end{quote}

\subsubsection{User Prompt Template}

For each question where both regions had intra-region agreement, the following prompt template was used:

\begin{quote}
\small
\textit{Compare already-identified concepts from two regions to find inter-regional agreement.}

\textbf{QUESTION:} \textit{[Question Text]}

\textbf{[REGION1] AGREED-UPON CONCEPTS (from intra-region analysis):}

\textit{1. Concept: `[Concept Name]'}

\textit{Evidence: `[Quote 1]', `[Quote 2]', `[Quote 3]'}

\textit{...}

\textbf{[REGION2] AGREED-UPON CONCEPTS (from intra-region analysis):}

\textit{1. Concept: `[Concept Name]'}

\textit{Evidence: `[Quote 1]', `[Quote 2]', `[Quote 3]'}

\textit{...}

\textbf{TASK:} \textit{Determine if any concept from [REGION1] matches any concept from [REGION2].}

\textbf{INTER-REGIONAL AGREEMENT CRITERIA:}
\begin{itemize}
    \item \textit{Agreement exists if ANY concept from [REGION1] is semantically similar to ANY concept from [REGION2]}
    \item \textit{For specific festivals and traditions, the festival or tradition names have to be an exact match for agreement}
    \item \textit{Both concepts must answer the same question}
    \item \textit{Semantic similarity includes synonyms, variations, and different ways of expressing the same idea}
\end{itemize}

\textbf{SEMANTIC MATCHING EXAMPLES:}
\begin{itemize}
    \item \textit{``emergency situations'' matches ``urgent circumstances'' (same underlying concept)}
    \item \textit{``cleaning house'' matches ``home tidying'' (same activity)}
    \item \textit{``touching feet'' matches ``feet touching'' (same gesture)}
    \item \textit{``festival sweets'' matches ``celebratory desserts'' (same food category)}
    \item \textit{``August'' matches ``august'' matches ``month of August'' (same month)}
\end{itemize}

\textbf{NO SEMANTIC MATCHING EXAMPLES:}
\begin{itemize}
    \item \textit{``pongal'' and ``lohri'' are both names of a harvest festival with the first for south and second for north but they are not semantically similar}
    \item \textit{``godh bharai'' and ``valaikappu'' are both names of a pregnancy ceremony with the first for north and second for south but they are not semantically similar}
\end{itemize}

\textbf{ANALYSIS PROCESS:}
\begin{enumerate}
    \item \textit{Compare each [REGION1] concept with each [REGION2] concept}
    \item \textit{Look for semantic similarity in the concept names and evidence quotes}
    \item \textit{If any pair matches, inter-regional agreement exists}
    \item \textit{If no concepts match, no inter-regional agreement}
\end{enumerate}
\end{quote}

\subsubsection{Required JSON Output Format}

The model was required to return structured JSON with the following schema:

\begin{quote}
\footnotesize
\begin{verbatim}
{
  "concept_comparisons": [
    {
      "region1_concept": "concept name from Region 1",
      "region2_concept": "concept name from Region 2",
      "semantic_match": true,
      "matching_explanation": "Why these concepts represent 
                               the same underlying idea"
    }
  ],
  "inter_regional_agreement": true/false,
  "matched_concepts": [
    {
      "unified_concept_name": "shared concept name",
      "region1_concept": "original concept from Region 1",
      "region2_concept": "original concept from Region 2",
      "semantic_explanation": "How these concepts are 
                               semantically similar"
    }
  ],
  "agreement_summary": "Brief explanation of whether and 
                        why inter-regional agreement was found"
}
\end{verbatim}
\end{quote}

\subsubsection{Example Application}

As an illustration, for a question about respectful gestures toward elders:

\begin{quote}
\footnotesize
\textbf{Question:} What is a common respectful gesture shown to elders in your region?

\textbf{South Region Agreed Concept:} ``Touching feet'' (4/5 responses mentioned variants: ``touch feet'', ``feet touching'', ``touching their feet'')

\textbf{North Region Agreed Concept:} ``Touching the feet'' (4/5 responses mentioned variants: ``we touch feet'', ``touching feet of elders'', ``feet touching'')

\textbf{Inter-Regional Analysis Result:} Agreement found

\textbf{Unified Concept:} ``Touching feet as a gesture of respect''

\textbf{Semantic Explanation:} Both regions independently converged on the same physical gesture (touching feet) as a sign of respect toward elders, with only minor variations in phrasing.
\end{quote}

\clearpage

\subsection{Universal Agreement Prompting Details}
\label{app:universal_agreement}

\subsubsection{Model Configuration}

\begin{itemize}
    \item \textbf{Model:} GPT-4o-2024-08-06
    \item \textbf{Temperature:} 0.1
    \item \textbf{Max Tokens:} 2000
\end{itemize}

\subsubsection{Question Matching Methodology}

Questions were matched between regions using normalized question text rather than question numbers. Text normalization involved:

\begin{itemize}
    \item Converting to lowercase
    \item Removing extra whitespace
    \item Stripping leading/trailing spaces
\end{itemize}

\subsubsection{System Prompt}

The following system prompt was used:

\begin{quote}
\small
\textit{You are an expert at comparing cultural concepts across multiple regions for semantic similarity. You must respond with valid JSON only.}
\end{quote}

\subsubsection{User Prompt Template}

For each question where all five regions had intra-region agreement, the following user prompt template was used:
\begin{quote}
\small
\textit{Compare already-identified concepts from 5 regions to find inter-regional agreement.}

\textbf{QUESTION:} \textit{[Question Text]}

\textbf{REGIONAL CONCEPTS:}

\textit{================================}

\textit{NORTH - AGREED-UPON CONCEPTS}

\textit{================================}

\textit{Concept 1: `[Concept Name]'}

\textit{Supporting Evidence:}

\textit{1. ``[Quote 1]''}

\textit{2. ``[Quote 2]''}

\textit{...}

\textit{================================}

\textit{SOUTH - AGREED-UPON CONCEPTS}
\textit{================================}

\textit{[Similar format for South region]}

\textit{[Similar sections for EAST, WEST, and CENTRAL]}

\textbf{TASK:} \textit{Determine if there is agreement across ANY or ALL of these regions: North, South, East, West, Central}

\textit{Systematically compare concepts across all 5 regions to determine:}
\begin{enumerate}
    \item \textit{Whether there is UNIVERSAL agreement (all 5 regions share the same concept)}
    \item \textit{Whether there is PARTIAL agreement (some but not all regions share concepts)}
    \item \textit{Whether there is NO agreement (each region has completely different concepts)}
\end{enumerate}

\textbf{INTER-REGIONAL AGREEMENT CRITERIA:}
\begin{itemize}
    \item \textit{Universal agreement exists if at least one concept from ALL regions (North, South, East, West, and Central) is semantically similar}
    \item \textit{Partial agreement exists if at least one concept is semantically similar for some and not ALL 5 regions}
    \item \textit{For specific festivals and traditions, names must be exact matches}
    \item \textit{All concepts must answer the same question}
    \item \textit{Semantic similarity includes synonyms, variations, and different expressions of the same idea}
\end{itemize}

\textbf{SEMANTIC MATCHING EXAMPLES:}
\begin{itemize}
    \item \textit{``emergency situations'' matches ``urgent circumstances'' (same underlying concept)}
    \item \textit{``cleaning house'' matches ``home tidying'' (same activity)}
    \item \textit{``touching feet'' matches ``feet touching'' (same gesture)}
    \item \textit{``festival sweets'' matches ``celebratory desserts'' (same food category)}
\end{itemize}

\textbf{NO SEMANTIC MATCHING EXAMPLES:}
\begin{itemize}
    \item \textit{``pongal'' and ``lohri'' are both harvest festivals but they are not semantically similar (different regional festivals)}
    \item \textit{``godh bharai'' and ``valaikappu'' are both pregnancy ceremonies but they are not semantically similar (different regional ceremonies)}
\end{itemize}

\textbf{NO UNIVERSAL AGREEMENT EXAMPLE:}
\begin{itemize}
    \item \textit{If North, South, West, and Central regions mention Diwali as the most popular festival but East mentions Durga Puja, that is not counted as universal agreement, instead it is partial agreement}
\end{itemize}

\textbf{ANALYSIS PROCESS:}

\textit{STEP 1: CREATE A COMPARISON MATRIX}
\begin{itemize}
    \item \textit{List all concepts from all 5 regions}
    \item \textit{For each unique concept group, identify which regions mention it}
    \item \textit{Example format:}
    
    \textit{Concept Group 1: ``Fasting/Observing Fast''}
    
    \textit{$\rightarrow$ Present in: North, South, West}
    
    \textit{Concept Group 2: ``Pongal''}
    
    \textit{$\rightarrow$ Present in: South only}
\end{itemize}

\textit{STEP 2: APPLY SEMANTIC MATCHING RULES}
\begin{itemize}
    \item \textit{Compare each concept from Region A with each concept from Regions B, C, D, E}
    \item \textit{Use the matching rules above to determine if concepts are semantically similar}
    \item \textit{Remember: Similar category $\neq$ Semantic match (e.g., both are festivals, but different festivals)}
\end{itemize}

\textit{STEP 3: IDENTIFY AGREEMENT PATTERNS}
\begin{itemize}
    \item \textbf{Universal Agreement:} \textit{At least ONE concept is shared by ALL 5 regions}
    \item \textbf{Partial Agreement:} \textit{At least ONE concept is shared by SOME regions (2 or more, but not all)}
    \item \textbf{No Agreement:} \textit{Each region has completely different concepts OR no semantic matches found}
\end{itemize}

\textit{STEP 4: DOCUMENT MATCHES}

\textit{For each matched concept group, clearly state:}
\begin{itemize}
    \item \textit{The unified concept name (the general term that encompasses all variations)}
    \item \textit{Which regions share this concept}
    \item \textit{What each region calls it (regional variations)}
    \item \textit{Why they are semantically similar (evidence from quotes)}
\end{itemize}
\end{quote}

\subsubsection{Required JSON Output Format}

The model was required to return structured JSON with the following schema:

\begin{quote}
\footnotesize
\begin{verbatim}
{
  "universal_agreement": true/false,
  "agreement_type": "universal|partial|none",
  "regions_in_agreement": ["region1", "region2", ...],
  "matched_concepts": [
    {
      "unified_concept_name": "shared concept name",
      "regions_sharing": ["region1", "region2", ...],
      "regional_variations": {
        "region1": "concept name in region1",
        "region2": "concept name in region2"
      },
      "semantic_explanation": "How these concepts are 
                               semantically similar"
    }
  ],
  "concept_matrix": [
    {
      "region1": "concept_name",
      "region2": "concept_name",
      "region3": "concept_name",
      "semantic_match": true/false,
      "explanation": "why they match or don't match"
    }
  ],
  "agreement_summary": "Brief explanation of inter-regional 
                        agreement patterns"
}
\end{verbatim}
\end{quote}

\subsubsection{Agreement Classification}

Universal agreement was classified into three mutually exclusive categories:

\begin{itemize}
    \item \textbf{Universal Agreement:} At least one concept is semantically similar across all 5 regions (North, South, East, West, Central)
    \item \textbf{Partial Agreement:} At least one concept is semantically similar across 2--4 regions, but not all 5
    \item \textbf{No Agreement:} No concepts are semantically similar across any subset of regions
\end{itemize}
\subsubsection{Example Application}

As an illustration, for a question about harvest festivals:

\begin{quote}
\footnotesize
\textbf{Question:} What is the main festival celebrated in your region?

\textbf{Regional Agreed Concepts:}
\begin{itemize}
    \item \textbf{North:} ``Diwali'' (4/5 responses)
    \item \textbf{South:} ``Diwali'' (4/5 responses)
    \item \textbf{East:} ``Durga Puja'' (4/5 responses)
    \item \textbf{West:} ``Ganesh Chaturti'' (4/5 responses)
    \item \textbf{Central:} ``Diwali'' (4/5 responses)
\end{itemize}

\textbf{Universal Agreement Analysis Result:} Partial agreement (not universal)

\textbf{Agreement Pattern:}
\begin{itemize}
    \item \textbf{Concept Group 1:} ``Diwali'' shared by North, South, and Central (3 regions)
\end{itemize}

\textbf{Classification:} Partial agreement (North, South, and Central share a concept, but not all 5 regions)

\textbf{Explanation:} While all regions celebrate festivals, the specific festival names are region-specific and not semantically similar according to the ``exact match for festival names'' rule. However, North, South, and Central share the same festival name (Diwali), resulting in partial agreement.
\end{quote}

\clearpage
\subsection{Dataset Structure}

\begin{minipage}\textwidth
\centering
\begin{tabular}{lccc}
\toprule
\textbf{Domain} & \textbf{Subcategories} & \textbf{Topics} & \textbf{Questions} \\
\midrule
Clothing and adornment & 2 & 4 & 59 \\
Communication & 2 & 4 & 47 \\
Education & 2 & 4 & 58 \\
Festivals and Rituals & 3 & 9 & 109 \\
Finance & 2 & 4 & 53 \\
Food processing and consumption & 2 & 4 & 50 \\
Interpersonal Relations & 3 & 6 & 86 \\
Traffic and transport behavior & 2 & 4 & 53 \\
\midrule
\textbf{Total} & \textbf{18} & \textbf{39} & \textbf{515} \\
\bottomrule
\end{tabular}
\captionof{table}{Question distribution across domains, subcategories, and topics}
\label{app_tab:coverage}
\end{minipage}

\clearpage

\subsection{Model Evaluation RASA}

\subsubsection{LLM-as-Judge Evaluation Details}
\label{app:llm_judge}

\paragraph{Model Configuration}

\begin{itemize}
    \item \textbf{Judge Model:} Google Gemini Flash 3.0 (via OpenRouter API\footnote{https://openrouter.ai/})
    \item \textbf{Temperature:} 0.0
\end{itemize}

\paragraph{System Prompt}

The following system prompt was used to establish the evaluation framework:

\begin{quote}
\small
\textit{You are an expert evaluator for cultural knowledge questions about India.}

\textit{Your task: Determine if a model's answer is factually correct compared to the gold standard answer.}

\textbf{Guidelines:}

\textit{1. CORRECT: The model's answer conveys ONLY the same core information as the gold answer}
\begin{itemize}
    \item \textit{Paraphrasing is acceptable}
    \item \textit{ALL details in the model's answer must appear in the gold answer}
    \item \textit{Equivalent terms are acceptable (e.g., ``clay stove'' = ``chulha'')}
    \item \textit{The answer should not contain significant additional facts, examples, or details beyond what the gold answer provides}
\end{itemize}

\textit{2. PARTIALLY\_CORRECT: The model's answer has some correct information from the gold answer but:}
\begin{itemize}
    \item \textit{Misses details from the gold answer}
    \item \textit{Contains additional information, facts, or examples that are not present in the gold answer (even if factually correct)}
    \item \textit{Provides extra context or details that go beyond the scope of the gold answer}
    \item \textit{Is too vague or incomplete}
\end{itemize}

\textit{3. INCORRECT: The model's answer:}
\begin{itemize}
    \item \textit{Contradicts the gold answer with wrong facts}
    \item \textit{Provides completely different information}
    \item \textit{Misses the main point entirely}
\end{itemize}

\textbf{Important:}
\begin{itemize}
    \item \textit{Focus on factual accuracy, not writing style}
    \item \textit{Consider cultural context and regional variations}
    \item \textit{Be strict about factual contradictions (e.g., ``jewelry'' $\neq$ ``cash'')}
    \item \textit{If the model adds information not in the gold answer (like additional examples, regional variations, or extra details), mark as PARTIALLY\_CORRECT even if the added information is accurate}
    \item \textit{The gold answer defines the scope - answers should not exceed that scope}
\end{itemize}
\end{quote}

\paragraph{User Prompt Template}

For each question-answer pair, the following template was used:

\begin{quote}
\small
\textbf{Question:} \textit{[Question Text]}

\textbf{Gold Standard Answer:} \textit{[Gold Answer]}

\textbf{Model's Answer:} \textit{[Predicted Answer]}

\textit{Evaluate the model's answer.}
\end{quote}

\paragraph{Required JSON Output Format}

The judge model was required to return structured JSON with the following schema:

\begin{quote}
\footnotesize
\begin{verbatim}
{
  "label": "CORRECT" | "PARTIALLY_CORRECT" | "INCORRECT",
  "reasoning": "brief explanation",
  "key_discrepancies": ["list any factual errors or 
                         significant additions"]
}
\end{verbatim}
\end{quote}

\subsubsection{Question Distribution}
\begin{table}[h]
\centering
\small
\begin{tabular}{lc}
\toprule
\textbf{Region} & \textbf{Number of Questions} \\
\midrule
North & 326 \\
South & 326 \\
East & 276 \\
West & 354 \\
Central & 348 \\
\midrule
\textbf{Total} & \textbf{1630} \\
\bottomrule
\end{tabular}
\caption{Distribution of Region-Anchored Short Answer questions by region}
\label{tab:rasa_distribution}
\end{table}

\subsubsection{RASA Sensitivity Analysis}
\begin{table}[h]
\centering
\small
\setlength{\tabcolsep}{6pt}
\begin{tabular}{lccc}
\toprule
\textbf{Model} & \textbf{w=0.3} & \textbf{w=0.5} & \textbf{w=0.7} \\
\midrule
Grok-4 Fast     & 39.4\% & 52.6\% & 65.7\% \\
GPT-5.2         & 38.9\% & 52.4\% & 65.9\% \\
Qwen3 VL        & 39.3\% & 51.6\% & 63.8\% \\
Llama 3.3 70B   & 37.5\% & 50.3\% & 63.2\% \\
Claude Sonnet 4.5 & 36.7\% & 51.7\% & 66.8\% \\
Mistral Large   & 36.7\% & 50.4\% & 64.1\% \\
Gemini 3.0      & 36.0\% & 51.1\% & 66.1\% \\
DeepSeek V3.2   & 35.7\% & 49.5\% & 63.4\% \\
\midrule
\textbf{Range}  & \textbf{3.7pp} & \textbf{3.1pp} & \textbf{3.6pp} \\
\bottomrule
\end{tabular}
\vspace{-2mm}
\caption{Sensitivity analysis: RASA overall accuracy under different partial credit weights. Model rankings remain stable across weights, with all models clustering within 3--4 percentage points at each weight setting. This confirms our findings are robust to the choice of partial credit weighting. Models are sorted by w=0.5 performance (our primary metric).}
\label{tab:rasa_sensitivity}
\end{table}

\subsubsection{Scoring Criteria}
\begin{minipage}\textwidth
\centering
\small
\setlength{\tabcolsep}{3pt}
\renewcommand{\arraystretch}{1.2}
\begin{tabular}{p{0.06\textwidth}p{0.24\textwidth}p{0.10\textwidth}p{0.24\textwidth}p{0.28\textwidth}}
\toprule
\textbf{Score} & \textbf{Question} & \textbf{Gold} & \textbf{Model Response} & \textbf{Rationale} \\
\midrule
\textbf{1.0} & In North India, during annual school functions or inter-school competitions, what transportation is typically arranged? & \textcolor{darkleafygreen}{Buses} & Schools typically arrange a \textcolor{darkleafygreen}{bus} for students' transportation & Correctly identifies the transportation arrangement \\
\midrule
\textbf{0.5} & In North India, do students usually remain in the same classroom while teachers rotate, or do they move between classes? & \textcolor{darkleafygreen}{Students remain} in the classroom and \textcolor{darkleafygreen}{teachers rotate} & \textcolor{darkleafygreen}{Students usually remain} in the same classroom while \textcolor{darkleafygreen}{teachers rotate}, \textcolor{red}{especially in primary and middle grades, though some secondary schools have students move for specialized subjects} & Correct core practice but adds content not in gold answer \\
\midrule
\textbf{0.0} & In South India, how do drivers typically behave toward pedestrians waiting at zebra crossings? & Drivers \textcolor{darkleafygreen}{wait for pedestrians to cross} & \textcolor{red}{Drivers often do not stop for pedestrians waiting at zebra crossings}, so pedestrians typically wait for a safe gap to cross & Contradicts the gold answer by describing opposite behavior \\
\bottomrule
\end{tabular}
\captionof{table}{Examples of the three-tier scoring system for RASA questions. \textcolor{darkleafygreen}{Green text} indicates correct information matching the gold answer; \textcolor{red}{red text} indicates incorrect or additional information.}
\label{app_tab:sa_scoring_examples}
\end{minipage}

\clearpage
\subsection{Model Evaluation RA-MCQ}
\subsubsection{Question Distribution}
\begin{table}[h]
\centering
\small
\begin{tabular}{lc}
\toprule
\textbf{Domain} & \textbf{Number of Questions} \\
\midrule
Interpersonal Relations & 16 \\
Education & 12 \\
Clothing \& Adornment & 8 \\
Food Processing & 9 \\
Communication & 5 \\
Finance & 9 \\
Festivals \& Rituals & 18 \\
Traffic \& Transport & 2 \\
\midrule
\textbf{Total} & \textbf{79} \\
\bottomrule
\end{tabular}
\caption{Distribution of Region-Agnostic Multiple Choice Questions by domain.}
\label{app_tab:ramcq_distribution}
\end{table}

\subsubsection{Chi-Square Test for Regional Selection Bias}
\label{app:chi_square_test}

We used a chi-square goodness-of-fit test to assess whether models exhibit regional selection bias in RA-MCQ questions.

\paragraph{Null Hypothesis}

The model selects uniformly at random from available options, with no regional preference.

\paragraph{Observed and Expected Counts}

\textbf{Observed count} $O_r$ for region $r$ is calculated by aggregating the model's actual selections:

\begin{equation}
    O_r = \sum_{q \in Q} \begin{cases}
\frac{1}{|R_{\text{selected}}(q)|} & \text{if } r \in R_{\text{selected}}(q) \\
0 & \text{otherwise}
\end{cases}
\end{equation}

where $Q$ is the set of all question instances (30 runs per unique question) and $R_{\text{selected}}(q)$ is the set of regions represented by the option the model selected in question instance $q$. If an option represents multiple regions, credit is split equally.

\textbf{Expected count} $E_r$ under uniform random selection:

For each question instance $q$ with $n_q$ options, if option $i$ represents region set $R_i$, each region receives:

\begin{equation}
\text{ExpectedCredit}_{r} = \frac{1}{n_q} \times \frac{1}{|R_i|}
\end{equation}

Total expected count for region $r$ across all instances:

\begin{equation}
E_r = \sum_{q \in Q} \sum_{i : r \in R_i} \frac{1}{n_q} \times \frac{1}{|R_i|}
\end{equation}

This accounts for: (1) varying numbers of options per question (3--5) and (2) multiple regions sharing the same option.

\paragraph{Example}

Two question instances:

\textbf{Instance 1 (3 options):} A$\rightarrow$North, B$\rightarrow$South, C$\rightarrow$\{East,West\}

\textit{Model selects Option A (North)}

Observed: North=1, South=0, East=0, West=0, Central=0

Expected: North=1/3, South=1/3, East=1/6, West=1/6, Central=0

\textbf{Instance 2 (5 options):} A$\rightarrow$North, B$\rightarrow$South, C$\rightarrow$East, D$\rightarrow$West, E$\rightarrow$Central

\textit{Model selects Option C (East)}

Observed: North=0, South=0, East=1, West=0, Central=0

Expected: Each region=1/5

\textbf{Aggregated across both instances:}

North: $O=1.0$, $E=0.533$; South: $O=0.0$, $E=0.533$; East: $O=1.0$, $E=0.367$; West: $O=0.0$, $E=0.367$; Central: $O=0.0$, $E=0.200$

\paragraph{Test Statistic}

The chi-square goodness-of-fit statistic is:
\begin{equation}
\chi^2 = \sum_{r} \frac{(O_r - E_r)^2}{E_r}, \quad df = 4
\end{equation}

where $df$ is the degrees of freedom (number of regions minus 1). Statistical significance was assessed at $\alpha = 0.05$.

\paragraph{Standardized Residuals}

To identify which specific regions deviate significantly from expectation, 
we calculate standardized residuals:
\begin{equation}
z_r = \frac{O_r - E_r}{\sqrt{E_r}}
\end{equation}

Values $|z_r| > 1.96$ indicate significant deviation at $\alpha=0.05$; 
$|z_r| > 2.58$ at $\alpha=0.01$; $|z_r| > 3.29$ at $\alpha=0.001$. 
Positive residuals indicate over-selection; negative indicate under-selection.
\clearpage

\subsection{Model Evaluation Prompts and Configuration}
\label{app:eval_prompts}

\subsubsection{Model Configuration}

\begin{itemize}
    \item \textbf{Temperature:} 1.0
    \item \textbf{Number of runs per question:} 30
    \item \textbf{API providers:} OpenAI API\footnote{https://platform.openai.com/docs/models} (for GPT models), OpenRouter\footnote{https://openrouter.ai/} API (for all other models)
\end{itemize}

\subsubsection{Region-Anchored Short Answer}

\textbf{Prompt Template:}

\begin{quote}
\small
\textit{You are answering a question about Indian cultural practices. Please provide a concise answer that directly answers the question.}

\textbf{Question:} \textit{[Question Text]}

\textbf{IMPORTANT:} \textit{Provide a direct answer in 1 sentence. Do not use conversational filler or any justifications. Simply answer the question in the most brief way possible.}

\textit{Answer:}
\end{quote}

\subsubsection{Region-Agnostic Multiple Choice (RA-MCQ)}

\textbf{User Prompt:}

\begin{quote}
\small
\textit{You are answering a question about Indian cultural practices. Please select the most appropriate answer from the given options.}

\textbf{Question:} \textit{[Question Text]}

\textbf{Options:}

\textit{A. [Option 1]}

\textit{B. [Option 2]}

\textit{C. [Option 3]}

\textit{...}

\textbf{IMPORTANT:} \textit{Respond with ONLY the letter of your chosen answer (A, B, C, D, or E). Do not provide any explanation or additional text.}

\textit{Your answer:}
\end{quote}

\clearpage
\subsection{Additional Tables, Figures, and Analyses}

\begin{minipage}\textwidth
\centering
\small
\begin{tabular}{p{3.2cm}p{1.2cm}p{1.4cm}p{2.8cm}p{4.5cm}}
\toprule
\textbf{Question} & \textbf{Model} & \textbf{Error Type} & \textbf{Gold Answer} & \textbf{Predicted Answer} \\
\midrule
In West India, what behaviors are commonly considered disrespectful while a teacher is speaking in class? & Llama & Over-explaining & Speaking to other students while a teacher is speaking is considered disrespectful. & In West India, behaviors such as \textcolor{green!70!black}{interrupting}, \textcolor{red}{making eye contact, or crossing arms and legs} are commonly considered disrespectful while a teacher is speaking in class. \\
\midrule
In West India, what religious holidays require people to avoid wearing or displaying certain colors? & Claude & Over-explaining & People typically avoid wearing black during auspicious religious holidays. & During \textcolor{red}{Amavasya (new moon day) and certain mourning periods,} wearing \textcolor{green!70!black}{black} is traditionally avoided, \textcolor{red}{while white is avoided during festivals, and during inauspicious days like Tuesdays and Saturdays some avoid black in certain communities.} \\
\midrule
In West India, how is a friend's parent usually greeted? Focus on the titles used to greet them. & Llama & Both & The standard greeting is \textcolor{orange}{Namaskar}, followed by \textcolor{orange}{``Uncle'' or ``Aunty,''} sometimes regional titles like ``Kaka/Kaku'' are used as well. & In West India, a friend's parent is usually greeted with titles such as \textcolor{green!70!black}{``Kaka''} or \textcolor{red}{``Kaki''} for the father and \textcolor{red}{``Aaji'' or ``Aai''} for the mother. \\
\midrule
In Central India, what ornaments are commonly worn by men on a daily basis? & Qwen & Over-explaining & Rings are most commonly worn by men on a daily basis. & Men in Central India commonly wear \textcolor{red}{a simple silver or brass chain, a wristwatch,} and sometimes \textcolor{green!70!black}{a ring} or \textcolor{red}{a thread (mangalsutra-style)} on a daily basis. \\
\bottomrule
\end{tabular}
\vspace{-3mm}
\captionof{table}{Examples of partially correct model responses on RASA with error type annotations. Over-explaining: correct core information but extraneous additions; Both: extraneous additions and key omissions; Underspecifying: lacks core information. \textcolor{green!70!black}{Green} = correct, \textcolor{red}{red} = erroneous/extraneous, \textcolor{orange}{orange} = omitted.}
\label{app_tab:partial_resp}
\end{minipage}

\begin{figure*}[!htbp]
    \centering
    \includegraphics[width=0.8\textwidth]{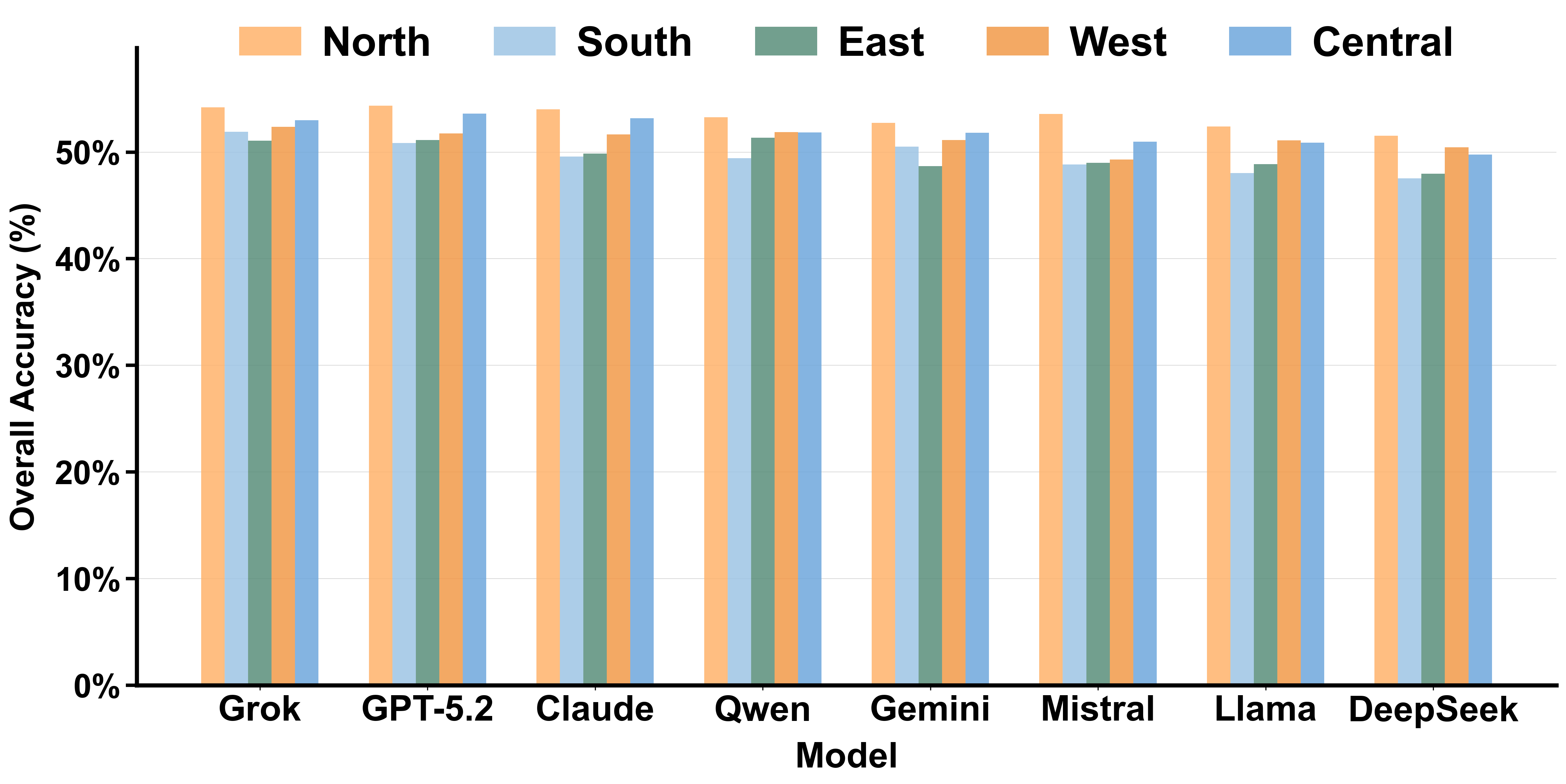}
    \caption{Overall Accuracy by regions of all models on RASA}
    \vspace{-3mm}
    \label{fig:rasa_regions_overall}
\end{figure*}

\begin{figure*}[!htbp]
    \centering
    \includegraphics[width=0.8\textwidth]{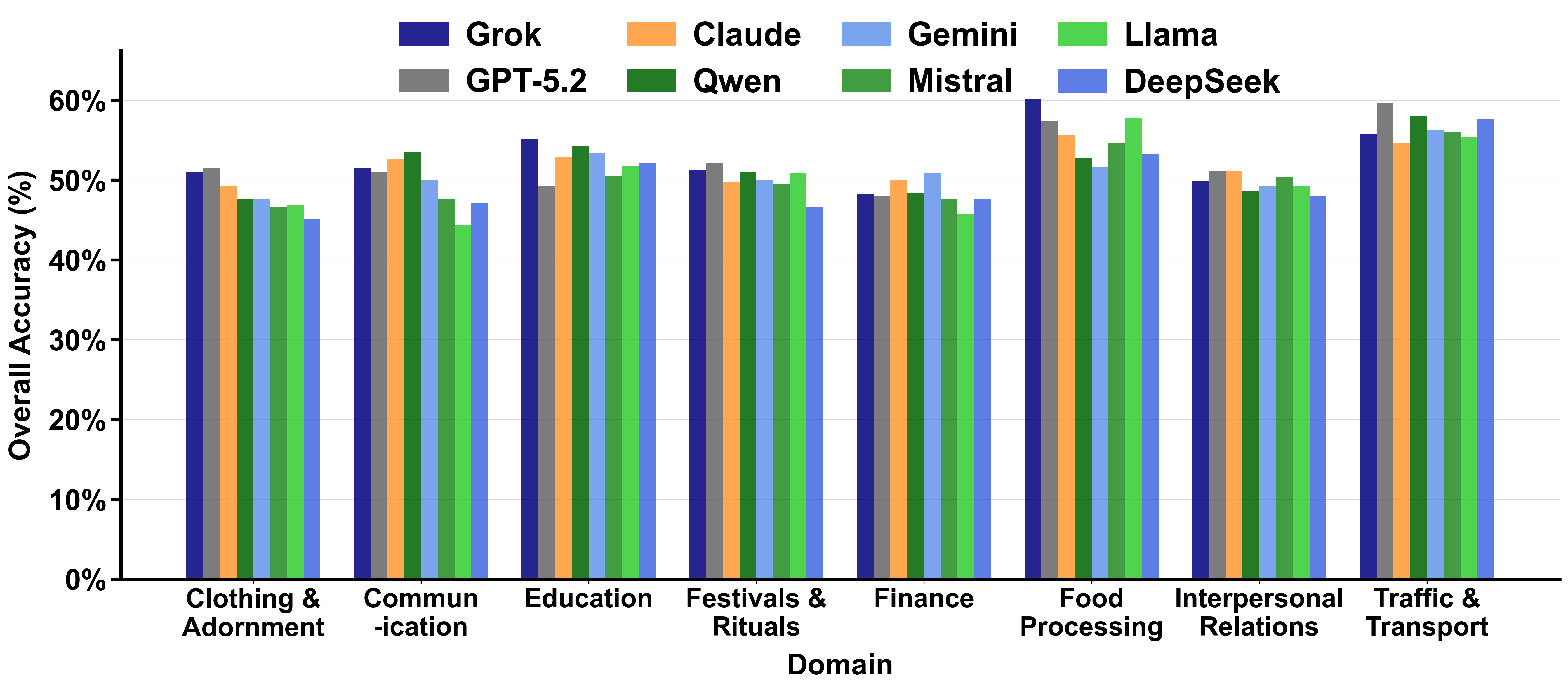}
    \caption{Overall Accuracy by domain of all models on RASA}
    \vspace{-3mm}
    \label{fig:rasa_domains_overall}
\end{figure*}

\begin{table*}[h]
\centering
\small
\setlength{\tabcolsep}{4pt}
\begin{tabular}{lcccc}
\toprule
\textbf{Model} & \textbf{$\chi^2$} & \textbf{df} & \textbf{p-value}  \\
\midrule
Qwen3 VL & 209.46 & 4 & <0.001***  \\
Gemini 3.0 Flash & 186.41 & 4 & <0.001***  \\
GPT-5.2 & 165.04 & 4 & <0.001***  \\
Llama 3.3 70B & 160.63 & 4 & <0.001***  \\
Grok-4 Fast & 143.32 & 4 & <0.001***  \\
Claude Sonnet 4.5 & 141.03 & 4 & <0.001***  \\
Mistral Large & 114.15 & 4 & <0.001*** \\
DeepSeek V3.2 & 80.70 & 4 & <0.001***  \\
\bottomrule
\end{tabular}
\vspace{-2mm}
\caption{Chi-square goodness-of-fit tests for regional selection bias in RA-MCQ. All models deviate significantly from uniform random selection (expected approx 20\% per region).}
\label{tab:ramcq_stats}
\end{table*}

\begin{table*}[!htbp]
\centering
\scriptsize
\setlength{\tabcolsep}{2.5pt}
\resizebox{\textwidth}{!}{
\begin{tabular}{lcccccccccc}
\toprule
& \multicolumn{5}{c}{\textbf{Observed (\% of Total)}} & \multicolumn{5}{c}{\textbf{Selection Ratio (Obs/Exp)}} \\
\cmidrule(lr){2-6} \cmidrule(lr){7-11}
\textbf{Model} & \textbf{North} & \textbf{South} & \textbf{East} & \textbf{West} & \textbf{Central} & \textbf{North} & \textbf{South} & \textbf{East} & \textbf{West} & \textbf{Central} \\
\midrule
Qwen3 VL        & 619 (26.1\%) & 394 (16.6\%) & 340 (14.3\%) & 358 (15.1\%) & 659 (27.8\%) & \textbf{1.32×} & 0.79× & 0.76× & 0.73× & \textbf{1.41×} \\
Gemini 3.0      & 531 (22.4\%) & 471 (19.9\%) & 379 (16.0\%) & 306 (12.9\%) & 683 (28.8\%) & \textbf{1.14×} & 0.95× & 0.84× & 0.63× & \textbf{1.46×} \\
GPT-5.2         & 563 (23.8\%) & 456 (19.2\%) & 356 (15.0\%) & 337 (14.2\%) & 658 (27.8\%) & \textbf{1.20×} & 0.92× & 0.79× & 0.69× & \textbf{1.40×} \\
Llama 3.3       & 566 (23.9\%) & 453 (19.1\%) & 357 (15.1\%) & 340 (14.3\%) & 652 (27.5\%) & \textbf{1.21×} & 0.91× & 0.80× & 0.70× & \textbf{1.39×} \\
Grok-4 Fast     & 589 (24.8\%) & 423 (17.9\%) & 315 (13.3\%) & 419 (17.7\%) & 624 (26.3\%) & \textbf{1.26×} & 0.85× & 0.70× & 0.86× & \textbf{1.33×} \\
Claude Sonnet   & 562 (23.7\%) & 442 (18.7\%) & 359 (15.1\%) & 363 (15.3\%) & 644 (27.2\%) & \textbf{1.20×} & 0.89× & 0.80× & 0.75× & \textbf{1.38×} \\
Mistral Large   & 575 (24.2\%) & 416 (17.5\%) & 389 (16.4\%) & 379 (16.0\%) & 612 (25.8\%) & \textbf{1.23×} & 0.84× & 0.86× & 0.78× & \textbf{1.31×} \\
DeepSeek V3.2   & 526 (22.4\%) & 450 (19.2\%) & 444 (18.9\%) & 346 (14.7\%) & 579 (24.7\%) & \textbf{1.14×} & 0.91× & 1.00× & 0.72× & \textbf{1.25×} \\
\midrule
\textbf{Average} & \textbf{561 (23.7\%)} & \textbf{438 (18.5\%)} & \textbf{360 (15.2\%)} & \textbf{356 (15.0\%)} & \textbf{639 (26.9\%)} & \textbf{1.21×} & \textbf{0.88×} & \textbf{0.82×} & \textbf{0.73×} & \textbf{1.37×} \\
\bottomrule
\end{tabular}}
\vspace{-2mm}
\caption{Regional selection in adversarial MCQs. Left: Observed counts and percentage 
of total selections (approx 2,370 per model). Right: Selection ratio (observed/expected), 
where expected counts derived from chi-square methodology. Ratio >1.0 = over-selection; 
<1.0 = under-selection. Bold indicates systematic over-selection across all models. 
Central India selected 1.37× expected rate (37\% over-selection); North India 1.21× 
(21\% over-selection); West India 0.73× (27\% under-selection).}
\label{tab:ramcq_detailed}
\end{table*}

\clearpage
\subsubsection{Regional Selection Details.} \label{app:regional_details}

\textbf{West India.} West India experiences the most severe under-selection across all models (12.9\%–17.7\%, 0.63–0.86× expected). Standardized residuals range from -3.1 (Grok) to -8.2 (Gemini), all significantly below expected rates. 

\textbf{East India.} East India shows similarly strong under-selection (13.3\%–18.9\%, 0.70–1.00× expected). Standardized residuals range from -1.9 (Mistral) to -5.7 (Grok). DeepSeek is a notable exception, selecting East India at near-expected rates (18.9\%, 1.00×, residual = 0.00), though it still under-selects West India.

\textbf{South India.} South India shows heterogeneous patterns across models (16.6\%–19.9\%, 0.79–0.95× expected). Three models significantly under-select: Qwen (16.6\%, 0.79×, residual = -4.7), Mistral (17.5\%, 0.84×, residual = -3.7), and Grok (17.9\%, 0.85×, residual = -3.3). Five models approach expected frequencies: Gemini (19.9\%, 0.95×, residual = -1.2), GPT-5.2 (19.2\%, 0.92×, residual = -1.9), DeepSeek (19.2\%, 0.91×, residual = -2.0), Llama (19.1\%, 0.91×, residual = -2.0), and Claude (18.7\%, 0.89×, residual = -2.5). Unlike the consistent over-selection of Central/North or under-selection of West/East, South India's treatment varies by model.

\clearpage

\end{document}